\documentclass[a4paper,fleqn]{cas-sc}

\usepackage[numbers]{natbib}
\usepackage{float}
\usepackage{caption}
\usepackage{cleveref}
\floatstyle{plaintop}
\restylefloat{table}
\usepackage{graphics}
\usepackage{rotating}
\usepackage{soul}
\usepackage{longtable}
\usepackage{multicol}
\usepackage{acronym}
\usepackage{subcaption}
\usepackage{mathptmx}
\usepackage{tabu}
\usepackage{graphicx}
\usepackage{fancyhdr}
\usepackage{mathtools} 
\usepackage{multirow}

\def\tsc#1{\csdef{#1}{\textsc{\lowercase{#1}}\xspace}}
\tsc{WGM}
\tsc{QE}
\tsc{EP}
\tsc{PMS}
\tsc{BEC}
\tsc{DE}

\begin{document}
\let\WriteBookmarks\relax
\def\floatpagepagefraction{1}
\def\textpagefraction{.001}
\shorttitle{}
\shortauthors{Ogunfowora et al.}

\title [mode = title]{Reinforcement and Deep Reinforcement Learning-based Solutions for Machine 
  Maintenance Planning, Scheduling Policies, and Optimization} 

\author[1]{Oluwaseyi Ogunfowora}[type=editor,
                        auid=000,bioid=1]
\ead{ogunfool@uvic.ca}
\credit{Conceptualization, Formal analysis, Investigation, Data Curation, Visualization, Writing - Original Draft, Writing - Review \& Editing}

\author[1,2]{Homayoun Najjaran}
\ead{najjaran@uvic.ca}
\credit{Writing - Review \& Editing, Supervision} 

\address[1]{Department of Mechanical Engineering, University of Victoria, Victoria BC, V8P 5C2, Canada}
\address[2]{Department of Electrical and Computer Engineering, University of Victoria, Victoria BC, V8P 5C2, Canada}
\cortext[cor1]{Corresponding author}

\begin{abstract}
Systems and machines undergo various failure modes that result in machine health degradation, so maintenance actions are required to restore them back to a state where they can perform their expected functions. Since maintenance tasks are inevitable, maintenance planning is essential to ensure the smooth operations of the production system and other industries at large. Maintenance planning is a decision-making problem that aims at developing optimum maintenance policies and plans that help reduces maintenance costs, extend asset life, maximize their availability, and ultimately ensure workplace safety. 
Reinforcement learning is a data-driven decision-making algorithm that has been increasingly applied to develop dynamic maintenance plans while leveraging the continuous information from condition monitoring of the system and machine states. By leveraging the condition monitoring data of systems and machines with reinforcement learning, smart maintenance planners can be developed, which is a precursor to achieving a smart factory. \\
This paper presents a literature review on the applications of reinforcement and deep reinforcement learning for maintenance planning and optimization problems. To capture the common ideas without losing touch with the uniqueness of each publication, taxonomies used to categorize the systems were developed, and reviewed publications were highlighted, classified, and summarized based on these taxonomies. Adopted methodologies, findings, and well-defined interpretations of the reviewed studies were summarized in graphical and tabular representations to maximize the utility of the work for both researchers and practitioners. This work also highlights the research gaps, key insights from the literature, and areas for future work.
\end{abstract}


\maketitle

\section{Introduction}
\label{The maintenance planning and Scheduling Problem}
\subsection{The maintenance planning and Scheduling Problem}
Maintenance activities take up 15\%-40\% of the total production costs in factories \cite{r1}. Machines/Assets undergo various failure modes that result in machine health degradation, affect system performance, and eventually cause machine failure. Degradation of machines whether under working or non-working conditions is inevitable and so is the need for maintenance. Maintenance, as defined by \cite{r2}, is a set of activities used to restore an item to a state in which it can perform its designated functions. Maintenance actions help to reduce machine failures, improve their reliability, and reduce the maintenance and production costs associated with unplanned downtime of machines. 
The maintenance planning and scheduling problem just like most planning problems is an optimization problem with the aim of developing efficient maintenance policies and adequately allocating maintenance resources and tasks to the geospatial problem. Maintenance plan optimization in literature is usually a multi-parameter optimization with the objective of deciding the right maintenance actions and when to perform the maintenance given a set of constrained or unconstrained maintenance capacities in a dynamic environment; ensuring machine availability, high yield, and low maintenance costs.\\
As earlier mentioned, maintenance of machines/assets is inevitable and that is why effective maintenance planning is paramount to ensure high asset availability with minimum cost. For most industries to remain competitive, they cannot afford the short or long-term costs and effects associated with inadequate planning of production and maintenance activities which can result in not meeting customer demands and loss of sales. Developing proper maintenance policies helps to reduce the costs associated with planned and unplanned downtime of machines and maintenance costs. Maintenance strategies are broadly classified into two categories: corrective maintenance (CM) and preventive maintenance (PM), the maintenance planning and optimization problem helps to find the optimum maintenance policies based on a chosen maintenance strategy.\\
Corrective maintenance can also be referred to as breakdown maintenance, this strategy adopts the run-to-failure approach; maintenance is only performed after machine failures, it is a reactive strategy. Usually, when machines fail under this policy, a replacement or major overhaul is required to get the machine back to good condition, and the costs and maintenance duration associated with corrective maintenance is usually very high. Asides from the costs associated with maintenance actions, a major drawback of the corrective maintenance strategy is that it does not avail the decision makers an opportunity to plan the maintenance actions. Machine failures happen abruptly and interrupt factory/running operations, which can cause significant losses due to unplanned downtime and require considerable high maintenance costs and resources for prompt maintenance actions. It can also cause accidents, for these reasons, the corrective maintenance policy is usually adopted for less critical equipment. \\ 
The preventive maintenance strategy, on the other hand, is to proactively shut down machines for maintenance to reduce failures and enhance the reliability of the machines. However, getting preventive maintenance policies that ensure smooth and efficient production is a non-trivial task because PM actions also impact the production system in ways that can incur additional production losses. The costs of excessive PM might outweigh its benefits, and inadequate PM can be ineffective in preventing unexpected machine failures. It is challenging to balance the delicate decision trade-offs that arise in PM for manufacturing systems, due to the complicated and nonlinear system dynamics that arise from interactions among machines\cite{r3}.
The preventive maintenance strategy can be divided into two main categories which are, scheduled and condition-based maintenance. \\
Scheduled maintenance is a preventive maintenance-based strategy where select maintenance activities are carried out at predefined intervals, scheduled maintenance activities are planned, and they are performed regardless of whether signs of deterioration or failures are prevalent or not. Even though this policy causes fewer abrupt failures, it can be very aggressive and eventually result in incurring unnecessary costs such as maintenance resource costs, and spare part costs for machines or assets that are still in good condition.  If the maintenance plans are not optimized, the costs associated with scheduled maintenance and production or industry-related costs from interventions will result in huge losses. Finally, scheduled maintenance cannot totally avoid abrupt failure due to stochastic, dynamic and nonlinear system dynamics resulting from interactions among machines which can cause machines to fail before the pre-defined maintenance period.\\
Condition-based maintenance also referred to as predictive maintenance suggests conducting maintenance actions based on some measurements of the system or component prior to failure. Predictive maintenance has been recognized as one of the most promising maintenance strategies for production systems because of its high efficiency and low cost compared to other strategies \cite{r4}. This approach helps to eliminate the unnecessary maintenance costs incurred in the scheduled maintenance approach while significantly reducing unscheduled breakdowns because the maintenance decisions are made based on the changing, real-time machine health conditions. Numerous works have been done in literature to harness the capabilities of machine-learning-based predictive models to accurately predict machine failures, make a diagnosis of the failure types, and prognostic efforts.  Other efforts to achieve better predictive results for machine learning-based predictive models due to their need for large amounts of data is in using generative models to generate more datasets. For instance, \cite{rm1} used a conditional generative adversarial network to generate faulty data from normal data using a few faulty samples. The aim of this work was to solve the problem of imbalanced datasets that engineers face when trying to use data-driven predictive models.  \\ 
While corrective and preventive maintenance actions are the two main maintenance strategies, many maintenance policies have been developed based on these strategies. The concept of machine maintenance optimization is not new, it is a concept that has been studied greatly in literature for so many years now, it dates to as far as the 1950s, and over the years, the maintenance optimization concept has had an unwavering goal of developing optimum plans and schedules for maintenance actions, it aims to find a balance between the costs and benefits of maintenance. To achieve this, efforts have been made in the following areas:
\begin{enumerate}
    \item \text Development of effective and universal maintenance policies such as age-dependent, periodic, failure-limit maintenance policies. \cite{r5} and \cite{r56} have presented review papers that summarize, classify, and compare the different maintenance policies for single and multi-unit systems. 
    \item \text Development of maintenance optimization models which are mathematical models with the aim of finding a balance between the costs and benefits of maintenance while taking into consideration available constraints \cite{r6}. 
    \item \text Joint or Integrated optimization: Maintenance actions affect other entities of production such as production scheduling, inventory, material handling, shift scheduling, and quality assurance, these connected, and sometimes conflicting entities are taken into consideration to develop optimum maintenance policies. e.g.,  \cite{r7} developed a control-limit-based maintenance policy for a serial production line while trying to create a balance or trade-off between the maintenance cost and product defective rate or yield of the machines.
    \item \text Ideally, there are maintenance planners that decide when and which machines are to be maintained based on the maintenance policies (decision-support systems) used in the production environment or industry, but with an increase in the number of machines or assets, the decision-making becomes too complex for the human planner to manage, hence, the need for an intelligent decision maker. Recent maintenance optimization researches use artificial intelligence to plan and schedule maintenance actions.\\ Also, as emphasized in \cite{r74}, there is a fast-growing demand for Intelligent Manufacturing Execution Systems (IMES) and one of the key functionalities of an Industry 4.0-ready MES is an intelligent maintenance planner and scheduler. 
\end{enumerate}
As mentioned above, the maintenance planning problem is an optimization problem, and several algorithms like the exact methods and meta-heuristics or global optimization algorithms have been used to solve it over the years. Reinforcement learning (RL) is a data-driven optimization algorithm that can be used to develop effective maintenance policies and there has been an upsurge in the application of RL to plan maintenance in the literature in recent years. This increase in the use of RL for maintenance planning is due to the increase in offline and real-time data from IoT devices and the high computing power that drives machine learning algorithms.
As earlier stated, predictive maintenance has gained the attention of many researchers in literature due to the need to avoid machine failures, and its ability to classify failures and accurately predict them, \cite{r8} summarises the predictive maintenance approaches. \\
Reinforcement learning application to the maintenance planning problem introduces an efficient and smooth transition between data-driven, condition-based maintenance predictive models and maintenance optimization models. Condition-based predictive models aim at reducing 
costs mainly by trying to predict when failure will occur and perform maintenance before that time, these models are specific to single-unit systems. The maintenance optimization methods generally try to minimize maintenance costs for multi-unit systems by developing optimum maintenance plans and schedules. Reinforcement learning methods as you will see in the coming chapters use a synergistic approach to combine the condition-based maintenance objective and the maintenance optimization objective(s) into a single problem formulation.
This opens opportunities to leverage condition-based maintenance (CBM) policies to optimize maintenance plans. Although CBM has been studied greatly in literature, maintenance plan optimization is still important because:
\begin{enumerate}
    \item \text The maintenance planning and optimization problem is focused on the relationship between costs and the maintenance decisions made. So, be it CBM or scheduled maintenance policies, a cost-benefit approach to planning maintenance actions helps to reduce overall maintenance costs. 
    \item \text Condition-monitoring systems focus on developing maintenance thresholds for individual components without taking into consideration the complex, non-linear interrelations due to interactions between machines or assets in a production environment. An efficient maintenance policy will incorporate all the machines into the maintenance planning decision and encourage sub-policies like group and opportunistic maintenance that take into consideration dependencies between sub-systems to take maintenance decisions thereby further reducing maintenance costs.
    \item \text Condition-based maintenance policies leverage their knowledge of the system’s health state or level to directly map the machine health states to maintenance actions. They can also afford to bypass the fault prognosis paradigm to accurately plan maintenance actions, allocate a set of maintenance tasks to resources, and ultimately reduce the system maintenance and production costs while ensuring the high availability of the machines.
\end{enumerate}

This work presents a literature review on reinforcement learning (RL), deep reinforcement learning (DRL), and hybrid methods which integrated RL and DRL with other methods to solve the maintenance optimization problem.\\
\textbf{Out-of-scope} publications are papers that used methods other than RL and its above-mentioned extensions to solve the maintenance optimization problem. Also, papers on RL/DRL-based maintenance planning for civil and structural infrastructures such as gas and water distribution systems; and bridge and railroad maintenance planning were not considered in this work.

\begin{figure}[h]
    \centering
    \includegraphics[width=0.7\textwidth]{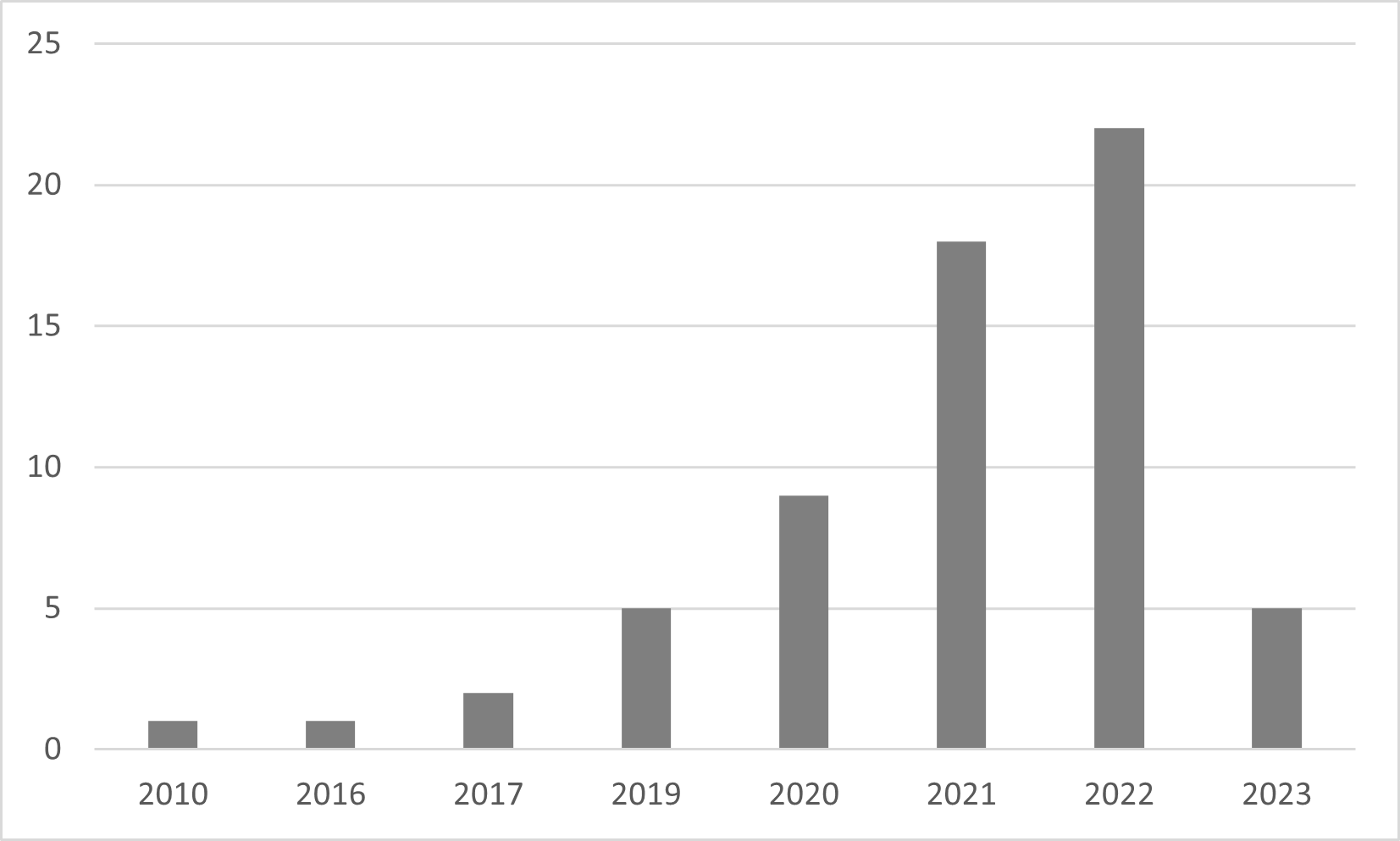}
    \caption{Relevant Publications/year}
    \label{Fig 1}
\end{figure}

Figure \ref{Fig 1} shows the number of RL and DRL-based maintenance planning and optimization publications per year over the past thirteen (13) years, the yearly analysis shows an evident upsurge in the use of RL and DRL for maintenance planning tasks between the years 2019 to 2023. There has been over an 80\% increase in the number of RL and DRL-based publications for maintenance planning in the literature.  

\subsection{Contributions of this work.}
The field of maintenance planning and optimization is not new, it has been explored greatly in literature, and maintenance is one of the main entities of an industry that needs planning amongst others which includes production planning. There are literature review papers that cover the applications of heuristics, meta-heuristics, RL, and deep RL for solving the production planning problem. \cite{r9} presented a review paper on the RL-based approach to manufacturing scheduling, \cite{r10} reviewed the state-of-the-art heuristics optimization algorithms used in solving the production scheduling problem,  \cite{r11} reviewed the applications of genetic programming for production scheduling, \cite{r12} presented literature reviews on meta-heuristics for dynamic and job-shop scheduling, and \cite{rm2} also presents a comprehensive review of Deep Reinforcement Learning for Machine Scheduling: Methodology, the State-of-The-Art, and Future Directions. However, in maintenance planning and optimization research, there are no literature review papers that specifically focus on the applications of exact methods, meta-heuristics, RL, and deep RL solutions for maintenance planning problems. These are all still open areas of research, and this work bridges the gap by presenting a literature review that focuses on the applications of reinforcement and deep reinforcement learning for maintenance planning and optimization.\\
This paper presents a systematic and integrative review, that focuses on methodologies, findings, and well-defined interpretations of the reviewed studies while finding common ideas and concepts, identifying methodological problems, and pointing out areas of research gaps. It also draws insights from existing literature and defines some areas of future work. 
This paper is structured in a way that it gives new researchers looking to apply RL or deep RL for maintenance planning and optimization a general overview and understanding of the underlying concepts, helping them to see the common, well-explored practices and approaches in the literature. It presents tables and figures that help them to make quick deductions and see relationships between predefined categories. This paper helps researchers understand the maintenance planning problem and  RL and DRL-based solutions. It also references other resources that can help to gain a deeper understanding of the core concepts such as different RL and DRL algorithms. \\
For experts and practitioners, this paper presents a quick glance through the literature, maps what has been achieved so far, highlights the problems that have been discussed and the existing solutions, and helps them to deduce new areas of research that can be explored in the future work section.\\
The RL solution is introduced and discussed in Section 2. Sections 3,4 contain a qualitative review, analysis, comparisons, collection of common ideas, and insights from the literature on the proposed RL and DRL-based maintenance planning solutions. Specifically, Section 3 discusses the maintenance planning problem formulation process. It summarizes and classifies reviewed publications in terms of the factors considered in the problem formulation process. It also contains figures and tables to compare and show relationships between different subsets of data within the domain of this literature review. \\
Section 4 reviews the formulation of the maintenance planning problem into the RL framework, state definitions, reward formulations, and environment. It also contains tables and figures that describe and show relationships between defined entities and discusses the RL, DRL, and hybrid RL and DRL-based algorithms that have been used in literature to solve the optimization problem. Sections 5 and 6 present key insights into the review analysis, implementation details and challenges, areas of future work, and conclusions respectively.\\

\begin{table}[htbp]
  \scriptsize
  \centering
  \caption{Table of Acronyms\label{tab 30}}
  \begin{tabular}{ l l l l }
  \toprule
    MDP & {Markov Decision Processes} & PPO & {Proximal Policy Approximation}\\
    RL &  {Reinforcement Learning} & DQN & {Deep Q Network}\\
    ML & {Machine Learning} & DPG & {Deterministic policy gradient}\\
    DRL & {Deep Reinforcement Learning} & VPG & {Vanilla Policy Gradient}\\
    CBM & {Condition-based maintenance} & PHM & {Prognostics and Health Management}\\
    PM & {Preventive Maintenance} & RUL & {Remaining Useful Life}\\
    CM & {Corrective Maintenance} & IOT & {Internet of Things}\\
    GA & {Genetic Algorithm} & GRL & {Reinforcement learning with Gaussian processes}\\
    GM & {Group Maintenance} & DDQN & {Double Deep Q Network}\\
    OM & {Opportunistic Maintenance} & TRPO & {Trust Region Policy Optimization}\\
    DP & {Dynamic programming} & DLQL & {Double-layer Q learning}\\
    2M1B & {Two-machine-one-buffer} & 5M4B & {Five-machine-four-buffer}\\
    DES & {Discrete-event simulation} & QMA & {Q learning with customized model-based acceleration}\\
    DPG & {Deterministic policy gradient} & DDPG & {Deep Deterministic policy gradient}\\
    TLD & {Transfer Learning with Demonstrations} & ELM & {Extreme Learning Machine}\\
    CPU & {Central processing unit} & DRQN & {Deep Recurrent Q-network}\\
    MARL & {Multi-agent Reinforcement Learning} & HMARL & {Hierarchical Coordinated Multi-agent Reinforcement Learning}\\
    QABC & {Artificial Bee Colony with Q learning} & DDMAC & {Deep Decentralized Multi-agent Actor Critic}\\ 
    LPRT & {Linear Programming Rollout} & MSSO & {Memetic Social Spider optimization algorithm}\\
    PERSEUS & {Randomized Point-based Value Iteration Algorithm} & PPO-LSTM & {Proximal Policy Approximation-Long short-term memory}\\
    SARSA & {State-Action-Reward-State-Action} & DRLSA & {Deep Reinforcement Learning Simulated Annealing}\\
    HDDE- RVNS & {Hybrid Genetic Algorithm – Random Variable Neighbour Search} & HGA-RVNS & {Hybrid Genetic Algorithm – Random Variable Neighbour Search}\\
  \bottomrule
  \end{tabular}%
\end{table}%

\newpage
\section{The Reinforcement and Deep Reinforcement Learning Solution}\label{sec:method}

\subsection{Reinforcement Learning}
Reinforcement Learning is one of the three (3) main paradigms of machine learning, the others being supervised and unsupervised learning. Contrary to the other paradigms; it adopts a trial-and-error method to take decisions. Reinforcement learning involves the RL agent exploring an unknown and uncertain environment to achieve a goal. The Markov state model forms a basis for the formulation of the RL paradigm that follows the notion that the available information about the current state is sufficient to predict the next state. It is based on the hypothesis that the accumulation of rewards through learning to take a sequence of optimal actions at every state in an environment is the maximization of the expected cumulative reward.
The formal framework for RL is from the problem of optimal control of Markov decision processes (MDP), the goal of the RL agent is to make an optimal decision based on the current Markov state.

\subsubsection{Inside a Reinforcement Learning system}
\begin{figure}[h]
    \centering
    \includegraphics[width=1\textwidth]{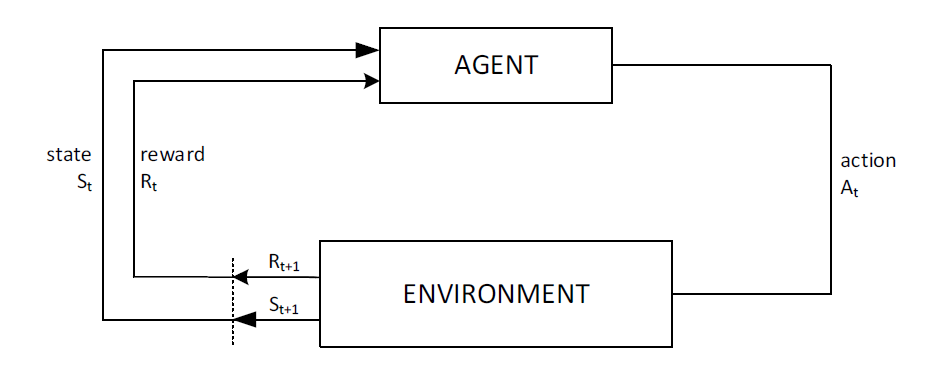}
    \caption{General RL Structure. Adapted from \cite{r15}}
    \label{Fig 2}
\end{figure}

The main elements of a reinforcement learning system are, the agent, the environment the agent interacts with, the policy that maps the agent’s states to actions and the reward an agent receives for taking certain actions. While the reward signal represents the immediate benefit of being in a certain state, the value function captures the cumulative reward that is expected to be collected from that state on, going into the future. The objective of an RL algorithm is to discover the action policy that maximizes the average value. figure \ref{Fig 2} shows the basic structure of the RL system.
The RL framework is originally based on MDP and the interaction between the environment and the agent is defined by this framework. A finite MDP problem can be defined by the tuple ($S$, $A$, $T$, $R$) here: $S$ is the set of states; $A$ is the set of possible actions; $T$ is the state transition probability and $R$, the reward function.  RL systems with a transition probability matrix that determines the next state are referred to as model-based RL algorithms while model-free frameworks do not build an explicit model of the environment. Asides from RL algorithms being categorized as model-based or model-based algorithms, they can also be broadly categorized as value-based, policy-based, and actor-critic algorithms. Value-based algorithms consider the optimal policy to be a direct result of estimating the value function of every state accurately, the value functions can be estimated using a recursive relationship as defined by the Bellman equation, popular value-based algorithms are state-action-reward-state-action (SARSA) and Q-learning. Policy-based algorithms, on the other hand, directly estimate the optimal policy without modeling the value function. By parameterizing the policy directly using learnable weights, they render the learning problem into an explicit optimization problem. Popular policy-based RL algorithms include the Monte Carlo policy gradient and deterministic policy gradient (DPG).
The most powerful RL algorithms are the actor-critic algorithms, they are a combination of the value-based and policy-based algorithms, both the policy (actor) and the value function (critic) are parameterized to enable effective use of training data with stable convergence. For detailed implementation information about these algorithms refer to  \cite{r15}.

\subsection{Deep Reinforcement Learning}
Despite the inherent sequential and dynamic nature of reinforcement learning algorithms, tabular RL algorithms are developed on static optimization formulations. As a result, many practical approaches are susceptible to optimality limitations, especially in problems with continuous state space, high-dimensional spaces, and long decision horizons, challenges such as the curse of dimensionality and the curse of history are faced. Tabular methods become impractical so deep RL methods are used to estimate the state values by using function approximators. Deep reinforcement learning applies a neural network to estimate the states instead of having to map every state to its values in a tabular form. It creates a more manageable solution space in the decision process. 

\subsection{The Reinforcement Learning or Deep Reinforcement Learning Solution and maintenance planning Problem}
In this section, we will discuss the reasons why reinforcement learning is a suitable solution for the maintenance planning problem and some of its advantages over other optimization methods such as heuristics and meta-heuristics.
\begin{enumerate}
    \item \text The maintenance planning problem can be dynamic, and stochastic; it is a fairly complex optimization problem because of the nature of the production environment or industry. The sequence of events or states is constantly changing, more specifically, the unit(s)/system degradation rates and profiles can change due to the complex, non-linear relationships between the environment entities such as the industry/factory operating conditions, seasonal influences, structural influences from other units in the system e.t.c. so they exhibit a stochastic behavior.\\ Because RL algorithms are developed based on the Markov Decision Process (MDP) framework, which is a stochastic process that models sequential decision-making in uncertain environments i.e.  MDPs can be used for planning in stochastic environments, RL can be to learn dynamic maintenance policies rather than static policies.

    \item \text Secondly, because the agent is allowed to interact with the environment during the decision-making process, the RL agents can optimize complex sequential decisions under uncertainties. Also, immediate maintenance actions taken in the maintenance planning problem may initially increase the cost of maintenance but subsequently lower the factory running cost, which is the case of delayed reward referred to in  \cite{r15} where actions may affect not only the immediate reward but also the next situations and through that all subsequent actions. The delayed-reward feature is one of the distinguishing features of RL asides from the ability to perform a trial-and-error search. \\ Another key feature of RL that makes it applicable to the maintenance planning problem is that contrary to other approaches that solve problems without considering how they fit in real-world or real-time decision-making, RL solves this problem from the onset through its interactive goal-seeking approach.  
    \item  More specifically, for condition-based maintenance policies, the degradation path of the machines and model of the production environment/industry and how the different entities relate to each other can be very difficult to quantify. RL permits the use of the model-free approach which can help to cope with non-linear complex systems. \\ Condition-based maintenance policies leverage their knowledge of the system’s health state or level to directly map the machine health states to maintenance actions, the machine health state can be represented by the degradation levels or remaining useful life and the process of modeling the degradation path or finding the remaining useful life is not non-trivial, it requires an in-depth understanding of maintenance analytics and if the health states are not correctly quantified and it is not a good representation of the actual machine state, this error propagates and influences the decision of any optimization method been used.  Another benefit of reinforcement learning methods over other optimization approaches is, we can harness the power of deep learning methods with RL and afford to bypass the fault prognosis or degradation modeling paradigm and directly map the states to optimal actions through learning from the environment. The agent learns the likelihood of failure of the machines at all states and learns the best time to perform maintenance. 
    \item \text The mathematical translation of the optimization objective to formulate the cost and constraint functions is the core of solving an optimization problem and it is a non-trivial task. A problem that is wrongly formulated will produce wrong policies which can have adverse effects when applied. We can however use the model-free RL approach and learn directly from the environment, the constraint functions need not be explicitly defined, the agent will learn it through experience, and this eliminates the burden and errors that arise from formulating the constraint functions. 

    \item \text Finally, if the RL agent is trained properly, it is bound to give better results and have a faster execution during inferencing compared to heuristics and meta-heuristics because it has explored every possible scenario and learned optimal maintenance actions for those scenarios.
    
\end{enumerate}

\section{Maintenance Planning and Optimization Model} \label{sec:biblio}
Condition monitoring can be leveraged to develop dynamic condition-based maintenance policies where the maintenance planning problem can be formulated as a Markov decision process with the goal of finding optimal dynamic maintenance actions that maximize the reward given the current state of the machine. It can be observed from RL and DRL-based publications for maintenance planning that the problem formulation follows a two-stage process. Initially, the maintenance planning problem is formulated as an optimization problem where, based on the system dynamics, available information about the system, resources, constraints, and the optimality criterion is defined. In the second stage, the maintenance planning problem is formulated as a Markov decision process with the goal of finding optimal maintenance policies. These two-stage problem formulations are discussed and summarized in sections 3 and 4 respectively.
\subsection{Problem Formulation}
\begin{figure}[t]
    \centering
    \includegraphics[width=1\textwidth]{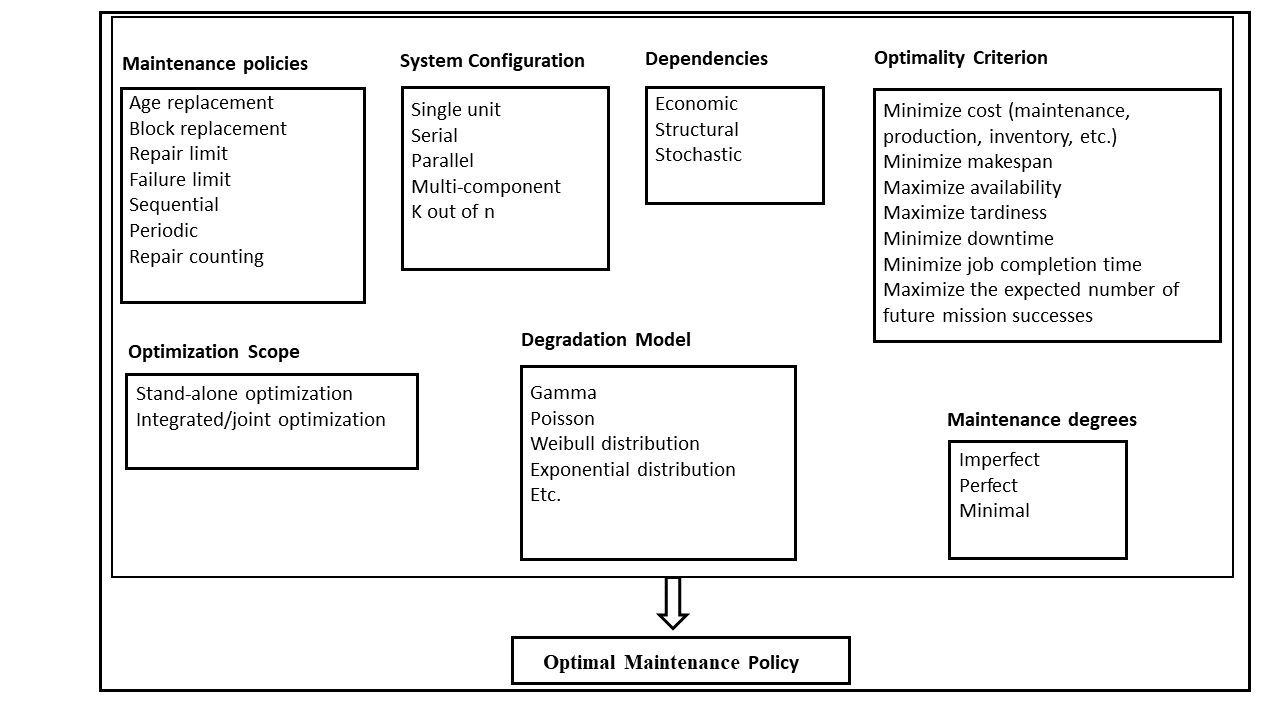}
    \caption{Factors considered in the maintenance planning and optimization problem formulation}
    \label{Fig 3}
\end{figure}

Just like most optimization problems, the process begins with problem formulation. The optimum design problem formulation is the translation of a descriptive statement of a design problem to a mathematical statement that can be optimized  \cite{r16}. It can be observed from reviewed publications that the system under consideration is first described. For the case of a maintenance planning problem, the system can be described based on many factors: a) \textbf{The maintenance policy} which is generally categorized as single-unit or multi-unit policies; single-unit and multi-unit policies can then be further divided into different sub-polices. b) \textbf{The dependencies} based on the dynamics of the system are in consideration. c) \textbf{The system configuration.} d) \textbf{The degradation model} that characterizes the failure of the machines or assets. e) \textbf{The maintenance degrees and effects.} f) \textbf{The optimality criterion.} g) \textbf{The optimization scope}, is categorized as joint, integrated, and stand-alone optimization in this paper.   \\
All the above-mentioned factors influence the way the maintenance planning problem is formulated, it affects the definition of the objective, and constraint functions. Due to a variety of factors that can be used to characterize the maintenance planning problem, the system description from one publication to the other varies greatly. To capture this diversity without losing touch with the uniqueness of each paper considered in this literature review, the taxonomies defined above are used to categorize the system, figure \ref{Fig 3} shows a breakdown of the factors that can affect the optimization problem formulation process. In the following sub-sections, RL and DRL-based maintenance planning policies are reviewed based on the taxonomies defined above. 

\subsection{The Maintenance Policies}
\begin{figure}
    \centering
    \includegraphics[width=0.7\textwidth]{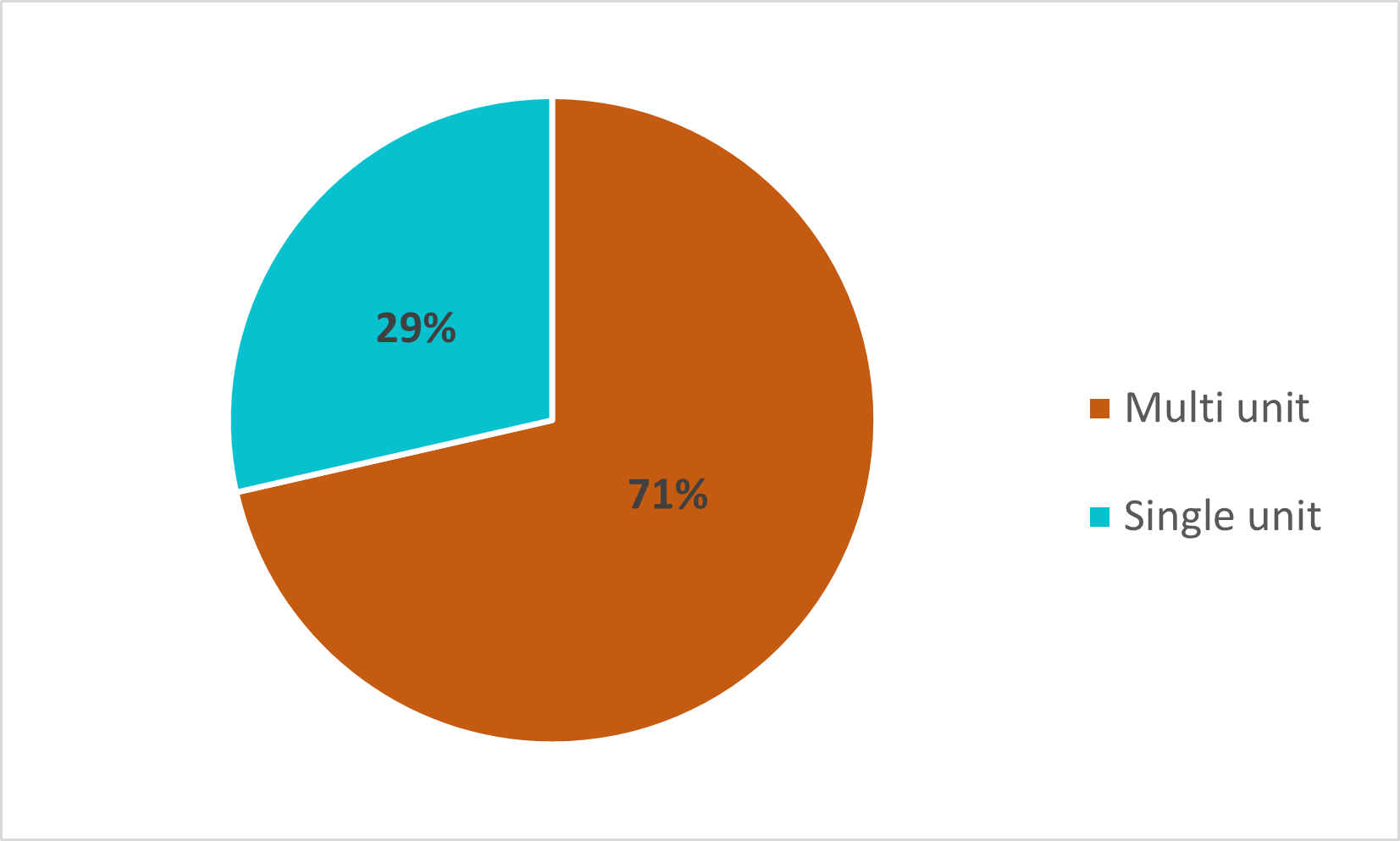}
    \caption{The ratio of RL and DRL-based single to multi-unit system policies in literature}
    \label{Fig 4}
\end{figure}

Machine maintenance policies have been studied under numerous application scenarios and are characterized by the target system. According to  \cite{r17}, all maintenance policies that have been studied in literature can be classified based on the structure of the target system which could be single-unit systems or multi-unit systems. Single-unit policies can be defined as maintenance policies developed for standalone systems where the relationship between system maintenance costs and maintenance decisions can be directly quantified. Single-unit systems can be further divided into six (6) sub-categories or policies which are: Age-dependent, Periodic, failure-limit policy, Sequential, Repair limit, Repair number counting, and Reference time policy. \\
Single-unit systems are well established in the literature, and they are used as a basis for the development of maintenance policies for multi-unit systems \cite{r17}. What differentiates single-unit systems from multi-unit systems is, multi-unit maintenance policies consider the dependencies between subsystems to develop maintenance plans or policies. Researchers make use of their understanding of the structure of the subsystems to formulate the objective function. Many multi-unit maintenance policies have been studied in literature and more specifically, it can be observed from the papers reviewed in this work that multi-unit policies have been considered more than single-unit policies. figure \ref{Fig 4} shows the ratio of single to multi-unit systems considered in RL and DRL-based solutions in the literature.
Similarly, multi-unit policies are divided into two main subcategories which are: Group and Opportunistic maintenance (GM and OM) policies, these policies can be combined or implemented separately. As earlier mentioned, single and multi-unit policies can be further grouped into various sub-policies as established in \cite{r17}. The following sub-sections summarize and classify maintenance policies for single-unit and multi-unit systems for the papers reviewed in this work.

\subsubsection{Maintenance policies of single-unit systems}
The maintenance planning problem is modeled as an MDP problem in order to use RL and DRL-based solutions to find the optimal policies. To achieve this, it is assumed that the state, degradation level, or health of the machine(s) or asset(s) are known by the RL agent. The machine or asset degradation levels can be represented by the age of the machines, constant or sequential periods, or a metric used to measure the reliability of the machine or asset. It can be observed from the literature that the failure-limit, age-based, periodic, and repair number counting policies are single-unit-based policies that have been used in RL and DRL-based solutions. \\ In this work, maintenance policies developed where the machine or asset degradation state is related to the component age are referred to as age-based policies. In this policy, corrective maintenance action is performed when the unit reaches a fixed age and fails. Age-based policies can be extended with the introduction of minimal repair(s) such that if the failure occurs before the predefined age, minimal repairs can be performed to return the unit to working conditions till it reaches the replacement age threshold. Preventive maintenance (PM) is proactively carried out to prevent machine failures. The following works \cite{r18}, \cite{r19}, \cite{r39}, \cite{r49}, \cite{r25}, \cite{r44} and \cite{r56} adopted the age-based policy in the development of their maintenance policies.\\
Periodic or time-based policy, here, units are preventively maintained at fixed time intervals. Only \cite{r23} and \cite{r52} adopted the periodic maintenance policy.
Failure-limit policies are the most used because condition-based maintenance strategies are prevalent among the papers that use RL and DRL solutions. Under this policy, preventive maintenance is performed before the system’s failure rate or reliability metric reaches a predefined level else corrective maintenance has to be done. This policy can also be extended by introducing minimal repairs such that if the failure occurs before the predefined level, minimal repairs can be performed to return the unit to working conditions. Every other reviewed publication aside from those that adopted the age-based and periodic policies all used the failure-limit policy. An extension of the failure-limit policy, which involves the combination of failure-limit policy and periodic policy was adopted in papers \cite{r50}, \cite{r49}, \cite{r4}, and \cite{r43}. 

\subsubsection{Maintenance policies for Multi-unit systems}
Multi-unit systems are made up of several single-unit systems. The group and opportunistic maintenance policies are specific to multi-unit systems. Group maintenance aims to establish policies that select a subset of systems that should be maintained at the same time while any unit in the subsystem is undergoing maintenance. By conducting multiple maintenance actions simultaneously, production losses and indirect maintenance costs can be reduced. As pointed out in \cite{r20}, the benefits of combining maintenance actions for multi-unit systems depend on spatial relationships between the units. If the components form a series configuration, less cost will be incurred if group maintenance is performed since the unavailability of a single unit causes the entire line or section to be unavailable anyway. This argument does not hold for parallel multi-unit systems, simultaneous maintenance is more desirable because GM policies would further reduce the system or network capacity and availability. \\ 
Following this definition, it can be observed that most authors that proposed a maintenance planning policy for serial production or flow lines considered the dependencies amongst the subsystems to develop efficient multi-unit policies. The authors of \cite{r18} and \cite{r56} developed preventive maintenance policies for serial production lines while taking into consideration the economic, structural, and stochastic dependencies amongst the subsystems. Authors of  \cite{r60}, \cite{r39}, \cite{r24}, and \cite{r21} considered only the structural and economic dependencies amongst the subsystems.\\
It can however be observed that without formally designing the optimization problem to perform group and/or opportunistic maintenance, a few publications have reported that the RL agents were able to learn GM and OM policies. \cite{r19} developed a CBM-based PM policy for a multi-state or degradation level serial production line by adopting a deep multi-agent, value-decomposition actor-critic DRL algorithm to obtain dynamic maintenance policies and reported that even though machines make PM decisions independently because each agent is assigned to individual machines on the production line, the agents learned to cooperatively perform OM and GM maintenance policies. The authors of \cite{r18} also proposed a centralized-single agent, DRL-based PM policy for serial production lines based on the double deep Q-network algorithm and reported that the agent learned GM and OM policies even though these concepts or rules were not explicitly provided during the learning process. The GM and OM policies were developed to leverage the dependencies such as economic dependency among multi-unit systems to make decisions and the RL agent interestingly was also able to learn these policies in its decision-making process just like a human maintenance planner will while keeping costs low. In \cite{r7}, \cite{r25}, \cite{r23}, and \cite{r73} even though the target systems had serial or flow line configurations, the authors did not consider the dependencies in their problem formulation so they were not able to capture the GM policies, these maintenance models can be extended to capture these dependencies in future works.
Asides from series or flow line systems, some multi-component systems can be structurally dependent. The authors of \cite{r31} developed a condition-based maintenance policy for multi-component systems and considered the economic and structural dependencies in the problem formulation.\\
Opportunistic maintenance (OM), the idea behind opportunistic maintenance policies is simply to take the opportunity of the downtime or repair of some system units to perform maintenance on other units with the aim of minimizing maintenance costs. The maintenance of a multi-component system differs from that of a single-unit system due to the inter-relations between the system units, system dependencies, competing failures, and stochastic failures; the maintenance of a unit can open an opportunity for the maintenance of other units. \cite{r26} considered this economic and stochastic dependency in the problem formulation of the maintenance policy for multi-component systems, under this policy during the maintenance of a given unit, based on the degradation states of other units, maintenance actions are carried out on the other units if required. \cite{r27} however, considered the opportunistic maintenance policy in a distinct way by proposing a predictive maintenance policy for parallel machines that aims at determining the time before a machine breaks down with low system load which corresponds to low opportunity costs rather than performing maintenance at the last possible time before breakdown. This approach was also adopted by \cite{r28} for the optimization of the operations and maintenance actions of wind turbines based on the availability of the maintenance crews. They reported that the policy learned by the proximal-policy-algorithm-based DRL agent outperformed the corrective, scheduled, and predictive maintenance strategies irrespective of the number of available maintenance crews because the agent learned to perform maintenance activities when the wind turbines are in a low power mode or demand is low while taking into consideration the health state of the wind turbine (anticipating failure) and the availability of the maintenance crews. It was able to outperform the predictive maintenance strategy which used the RUL threshold to determine when maintenance actions should be carried out because rather than just performing maintenance when the RUL threshold is reached, it took actions when the opportunistic costs are lower.\\
Finally, serial production lines with buffers in between the machines can be classified under multi-unit systems and the above-mentioned sub-policies should also apply to them, \cite{r18}, however, established the fact that these policies only apply to close-interconnected serial production lines (i.e., serial production lines with no buffers in between them) and does not apply to serial production lines with buffers in between them because the GM and OM policies are developed under the idea that when one machine is under maintenance, the others can receive maintenance simultaneously without incurring extra production loss through unavailability of machines or assets. However, this does not hold for serial production lines with intermediate buffers because the buffers in between machines could delay the propagation of machine stoppage from the maintained machine to the adjacent machines \cite{r29} and \cite{r30}. So, to adequately evaluate the system performance and develop a true objective function for serial production line with intermediate buffers, the system dynamics need to be considered more carefully because, if some key variables such as buffer levels are not included in the problem formulation, the learned policies would not reflect the real system dynamics. The authors of \cite{r18} tackled this problem through a prior understanding of the system dynamics based on a data-driven analytical model for serial production lines that they developed in their previous work. The data-driven model can efficiently evaluate the real-time dynamic behavior of the production system and the system production loss can be adequately evaluated.

\subsection{Dependencies}
Typically, there are three (3) categories of dependencies. Economic, stochastic, and structural dependencies, these dependencies are common with multi-unit systems and they are the basis for the above-mentioned multi-unit sub-policies. Economic dependence allows for shared indirect maintenance costs such as set-up costs and downtime costs when multiple components are maintained together. Structural dependence refers to when several sub-components physically or structurally form a complete part or unit, in cases like that, less cost is incurred when the components are maintained together. It also enhances the reliability of the machine by eliminating frequent interventions on the system. For instance, \cite{r26} extended structural dependence to the concept of competing risks where a system fails if any of its components fails, e.g., a modern computer could fail due to the failure of its CPU, storage unit, or operating system, whichever occurs first. The competing risks also impose economic dependence among components since the downtime of the system, after one component fails, is shared by all the components. for instance, the development of the PM policy.\\
Stochastic dependence occurs when the degradation of components has correlations and interactions, if these components are known, then they are better maintained together. 

\subsection{System Configuration}
\begin{figure}
    \centering
    \includegraphics[width=0.7\textwidth]{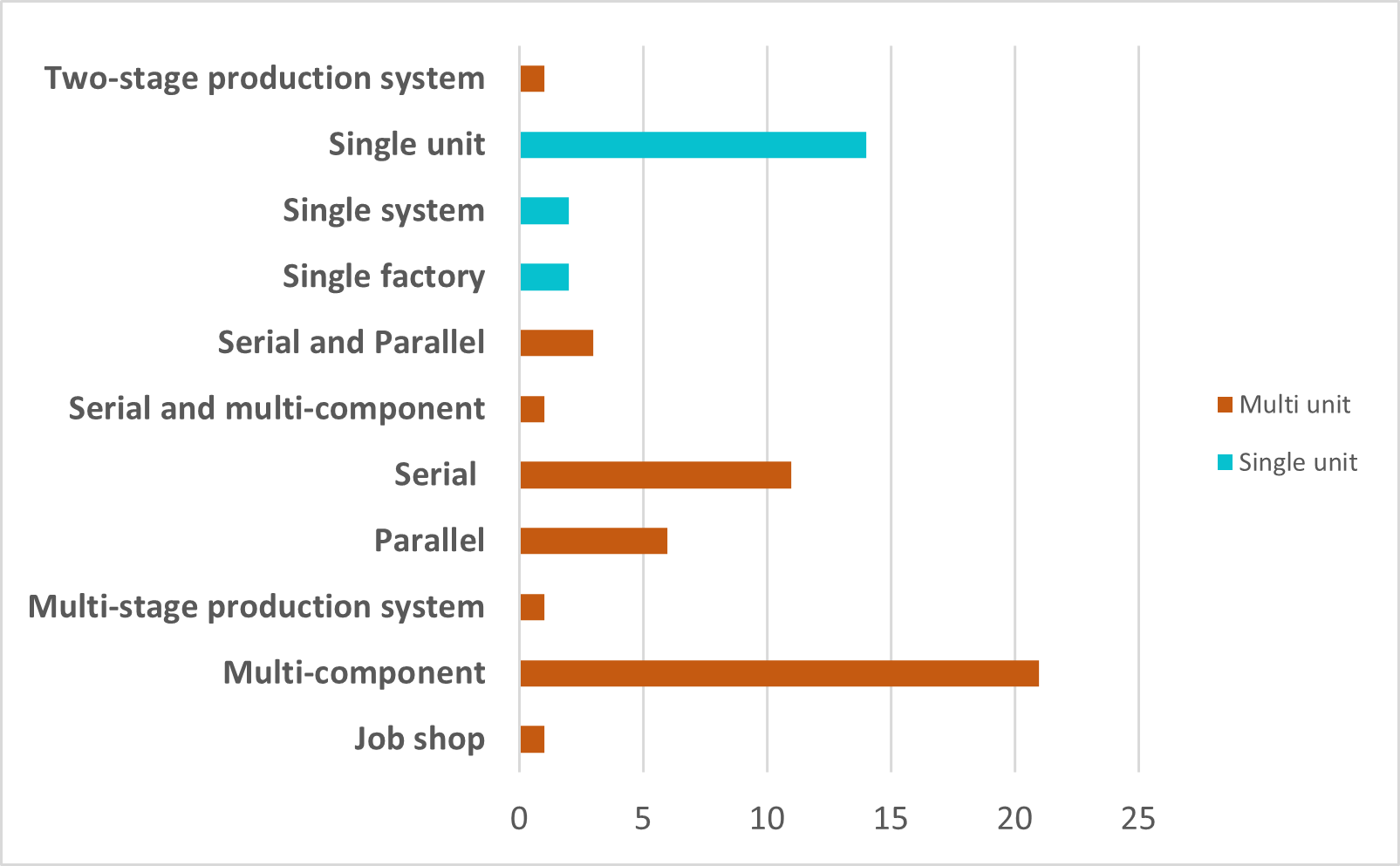}
    \caption{System configurations considered in the reviewed publication}
    \label{fig 5}
\end{figure}

This section groups the papers within the scope of this work based on the system configuration.
System configurations refer to how the machines or components are connected with each other to achieve predefined tasks. The five (5) main system configurations seen in reviewed publications are the single-unit, parallel, serial, multi-component, and multi-stage system configurations.\\
The single-unit environment refers to a single machine or component and it is the basic building block for other configurations. The parallel machine configuration is a collection of multiple machines that can work in parallel, the defined tasks can be processed on any of the machines at the same time. Serial systems have machines connected in series, the serial configuration requires tasks to be carried out in a sequential manner. Multi-component configurations in the reviewed papers are used to refer to a system of sub-components that make up a machine / a system network, these components can be connected either in parallel, in series, or a combination of both. The multi-stage production system is very similar to the multi-components system in terms of configuration but it is usually used to describe a system of machines connected together in production to process jobs.
In \ref{fig 5}, you will observe that we have the single system and single factory configurations in the figure as well this is because the entire sub-unit(s) or components of a single machine or machines in a factory can be modeled as a single-unit system. Also, the serial configurations for production lines can include the closed-interconnected lines or flow lines which is a collection of machines arranged in sequence such that jobs can be passed from one machine to the other with no intermediate buffers between the machines and serial production line with intermediate buffers.
From figure \ref{fig 5} it is obvious that multi-component systems followed by a single unit and serial production lines are the most studied system configurations in RL and DRL-based maintenance planning policies in literature. Table \ref{tab 31} classifies the reviewed publications in terms of the system configurations and optimization scope.

\renewcommand{\arraystretch}{1.5}
\begin{table}[htbp]
  \scriptsize
  \centering
  \caption{Reviewed publications grouped in terms of system configuration and optimization scope}
    \begin{tabular}{ p{3.2cm}|p{3.2cm} p{3.2cm} p{3.2cm}  }
    \toprule
    \multicolumn{4}{c}{Optimization Scope} \\
    \midrule
    \multicolumn{1}{l}{System Configuration} & Stand-alone  & Integrated Optimization & Joint Optimization \\
    \midrule
    Single unit/factory     & \cite{r57}, \cite{r41}, \cite{r34}, \cite{r37}, \cite{r42}, \cite{r65}, \cite{r66}, \cite{r71}, \cite{rn4} & \cite{r53}, \cite{r45}, \cite{rn12} & \cite{r52}, \cite{r49}, \cite{r4}, \cite{r68}, \cite{rn2}, \cite{rn6} \\
    Multi-component     & \cite{r31}, \cite{r26}, \cite{r35}, \cite{r3}, \cite{r28}, \cite{r58}, \cite{r59}, \cite{r43}, \cite{r62},
    \cite{r46}, \cite{r44}, \cite{r63}, \cite{r64}, \cite{r67}, \cite{r70}, \cite{rn1} , \cite{rn5}, \cite{rn10}, \cite{rn11} & \cite{r51}, \cite{r47} & - \\
    Parallel Configuration     & \cite{r27}, \cite{r40}, \cite{r61}, \cite{rn3},\cite{rn8} & - & \cite{rn13}\\
    Serial Configuration     & \cite{r18}, \cite{r19}, \cite{r21}, \cite{r24}, \cite{r56}, \cite{r73} & \cite{r7}, \cite{r22}, \cite{r25}, \cite{rn9} & \cite{r23}  \\
    Serial and Parallel     & \cite{r39}, \cite{r60}  & - & - \\
    Serial and multi-component     &  - & \cite{rn7} & - \\
    Two and multi-stage configuration     &  \cite{r72} & - & \cite{r69} \\
    Job shop     &  -  & - & \cite{r50}\\
    \bottomrule
    \end{tabular}%
  \label{tab 31}%
\end{table}%

\subsection{Degradation Model}
Machines or assets in use or out of use are subject to degradation and that is why maintenance activities are required to return them to functional states. Degradation models capture the life of the machines or assets; choosing a proper degradation model is one of the major steps in the maintenance modeling of RL and DRL-based maintenance planning policies because the dynamic degradation states of the machines or assets, specific to the problem in consideration are continually inspected and used to take decisions. Be it a model-based or model-free RL approach or scheduled and CBM-based policies, the degradation model is required. The maintenance problem can also be modeled in a way that the maintenance effect and maintenance costs are directly related to the degradation states of the machine, for instance, in \cite{r32}, the perfect maintenance or preventive replacement action returns the equipment to a brand-new state while the imperfect maintenance action improves the degradation state of the component but accelerates the degradation rates. \\
The degradation model shows how the systems or corresponding sub-systems degrade over time.  In literature, due to a lack of failure information and data on the systems or the difficulty in collecting this information, degradation modeling techniques such as the Gamma process, and Poisson processes, have been used more frequently to model degradation. To model stochastic deterioration, failure rate functions or stochastic processes such as random deterioration rates, Markov processes, or non-decreasing jump processes which are the case of Gamma can be used \cite{r33}.\\ According to \cite{r33}, the gamma process is used to formulate a monotonically increasing degradation path and it can be used to model degradation processes such as wear, corrosion, and creep in systems, the Poisson process is a continuous-time process in which a random shock of random magnitudes arrives at the model or system at random times. 
The Gamma, Poisson, Weibull, and Weiner processes are the most used degradation models in RL and DRL-based maintenance models in the literature. \cite{r24}, \cite{r34}, and \cite{r7} adopted the Gamma process to model machine degradation, \cite{r26}, \cite{r21}, and \cite{r35} used a combination of the gamma and Poisson process to model the degradation path of the components. In the publications that used a combination of the gamma and Poisson process to model the degradation path of the components, during the operational stage, each component in the system undergoes individual or component-level degradation which follows a Gamma process as well as a system-level degradation which follows the Poisson distribution and comes into the system in form of random shocks. The combination of these degradation processes forms the cumulative degradation of each component at a time. It can be observed that the papers that adopted the combination of Gamma and Poisson degradation models were developed for multi-component systems and this combination was used to capture the stochastic deterioration process that is observed in multi-component systems which also applies to many real-life applications, for instance, according to \cite{r26}, a real-life example of a stochastic degradation process would be in the case of the gearbox of an engine that contains many gears, the individual gears in the gearbox are subject to component-level degradation but the start and stop actions of the engine can impose a system level shock.\\ \cite{r37} also used the Poisson process to model the degradation paths.
The Weibull distribution was first introduced in 1939 by Waloddi Weibull. According to \cite{r38}, it is a statistical distribution that can be used to describe observed system failures. \cite{r19}, \cite{r27}, \cite{r39}, and \cite{r40} modeled their degradation path following the Weibull distribution. \cite{r41} and \cite{r42} adopted the Weiner process to model the system degradation paths. Another way the degradation paths of machines or assets have been modeled in literature is under the assumption of the availability of a prognostics model that can tell the remaining useful life or the time to failure of components, \cite{r31}, \cite{r59}, \cite{r43}, \cite{r44} and \cite{r45} all modeled their degradation based on the prognostics capabilities of the system. 
Other distributions such as the exponential, Gaussian, uniform, and linear distributions have been used in literature to model the degradation process. \\  An observation from CBM-based RL and DRL-solutions in the literature however is that the chosen degradation model parameters are usually assumed due to a lack of failure data, only a few publications like papers \cite{r46}, \cite{r47} and \cite{r64} generated the equipment degradation behaviors from data.\\
Discretized state degradation or the Markov process has also been used to represent the degradation paths of machines or assets in literature. Most of the publications that use the discretized Markov state process however do it under the assumption that they know the degradation levels of the machines and can adequately tell when the machine(s) reaches this degradation level,  In cases like this, the authors pay more attention to the RL algorithm optimization problem and give less attention to the degradation modeling part. Some authors such as \cite{rn1}, \cite{rn2}, \cite{rn4}, and \cite{rn13} did not just assume the existence of a prognostics model or a remaining-useful life predictor, they developed an end-to-end data-driven machine learning-based maintenance planners. They used various deep-learning architectures such as convolutional neural networks to first develop a predictive model to estimate the remaining useful life and used this information to represent the states of the machine. 

\subsection{Maintenance degrees}
While some authors simply consider the maintenance actions as perfect under the idea that the preventive maintenance action returns the machine state to as-good-as-new, most authors consider different degrees of maintenance that can either return the machine or assets to an as-good-as-new state, as-bad-as-old or other states depending on the way the maintenance effects were modeled. This factor is specific to the industry/factory maintenance approach, machine, equipment, or component in consideration. 
\subsection{Optimality Objective and Criterion}
\begin{figure}
    \centering
    \includegraphics[width=0.7\textwidth]{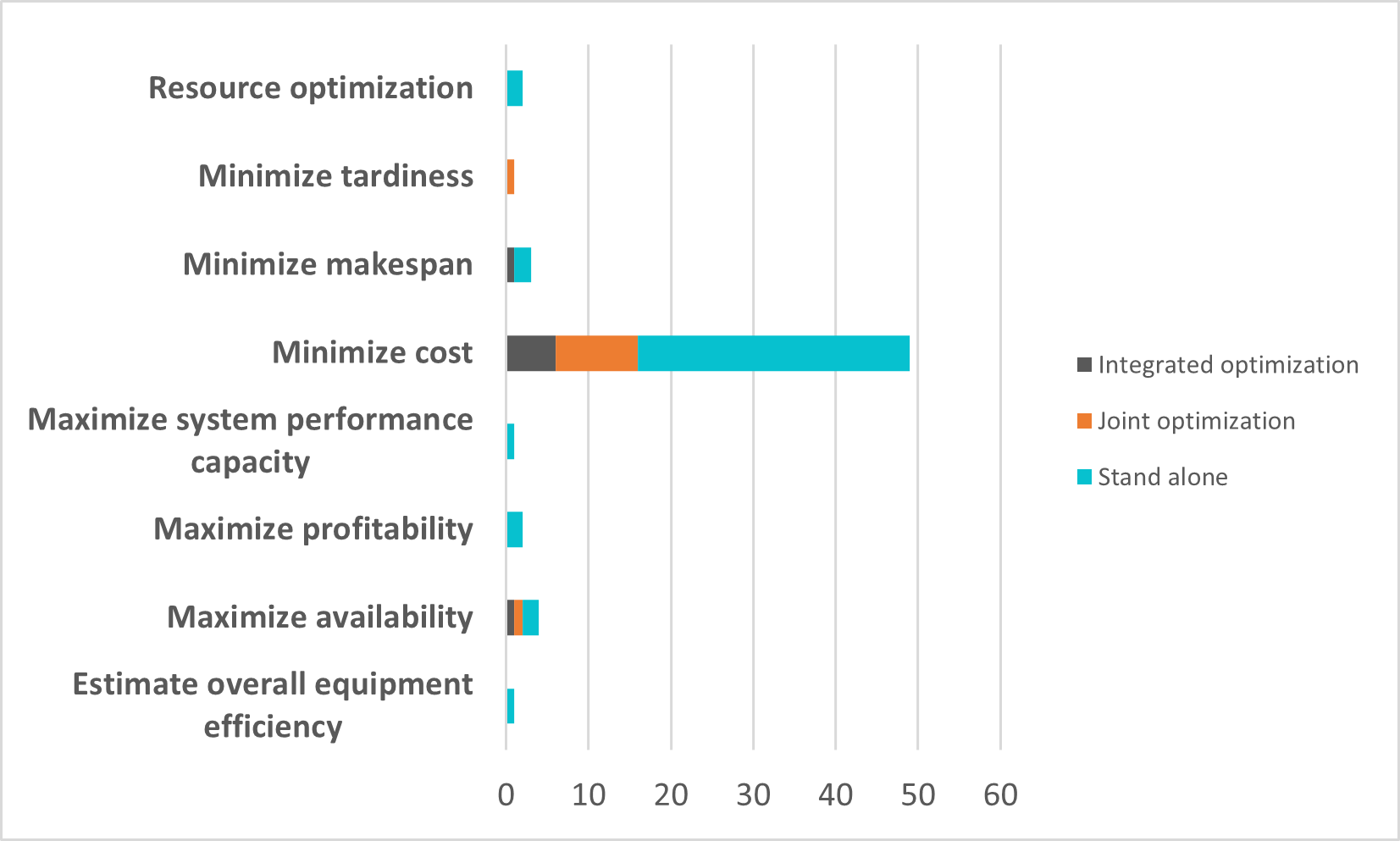}
    \caption{Optimality Criterion and Optimization scope}
    \label{fig 6}
\end{figure}

The maintenance planning problem is usually a multi-objective optimization problem; the main objectives are to develop maintenance plans or take maintenance actions that minimize the overall system or maintenance costs and maximize machine reliability and availability.  Even though these are the primary objectives of the maintenance optimization problem, other objectives like maintenance resource allocation and optimization have also been considered in the literature. The authors of \cite{r46}, and \cite{r44} developed an RL-based maintenance policy with a multi-objective of finding a balance between minimizing maintenance costs and efficiently allocating multiple maintenance crews in industrial IOT devices and wind farms respectively.\\
While the optimization objectives can vary from one problem or application to another, the optimality criterion is the measure of goodness of the system in question, this is the function that is optimized and generally, the maintenance multi-objective optimization problems can be formulated into a single function referred to as the optimality criteria. From publications reviewed in this work and most maintenance optimization problems, the most widely used optimality criterion is (a) minimize system or maintenance costs (b) maximize system availability and reliability (c) minimize tardiness (d) minimize makespan. (c) and (d) are usually considered when joint optimization of production and maintenance actions are conducted. From figure \ref{fig 6} it can be observed that the most used optimality criteria in the reviewed papers are to minimize total system cost or maintenance costs.
\subsection{Optimization scope}
\begin{figure}
    \centering
    \includegraphics[width=0.7\textwidth]{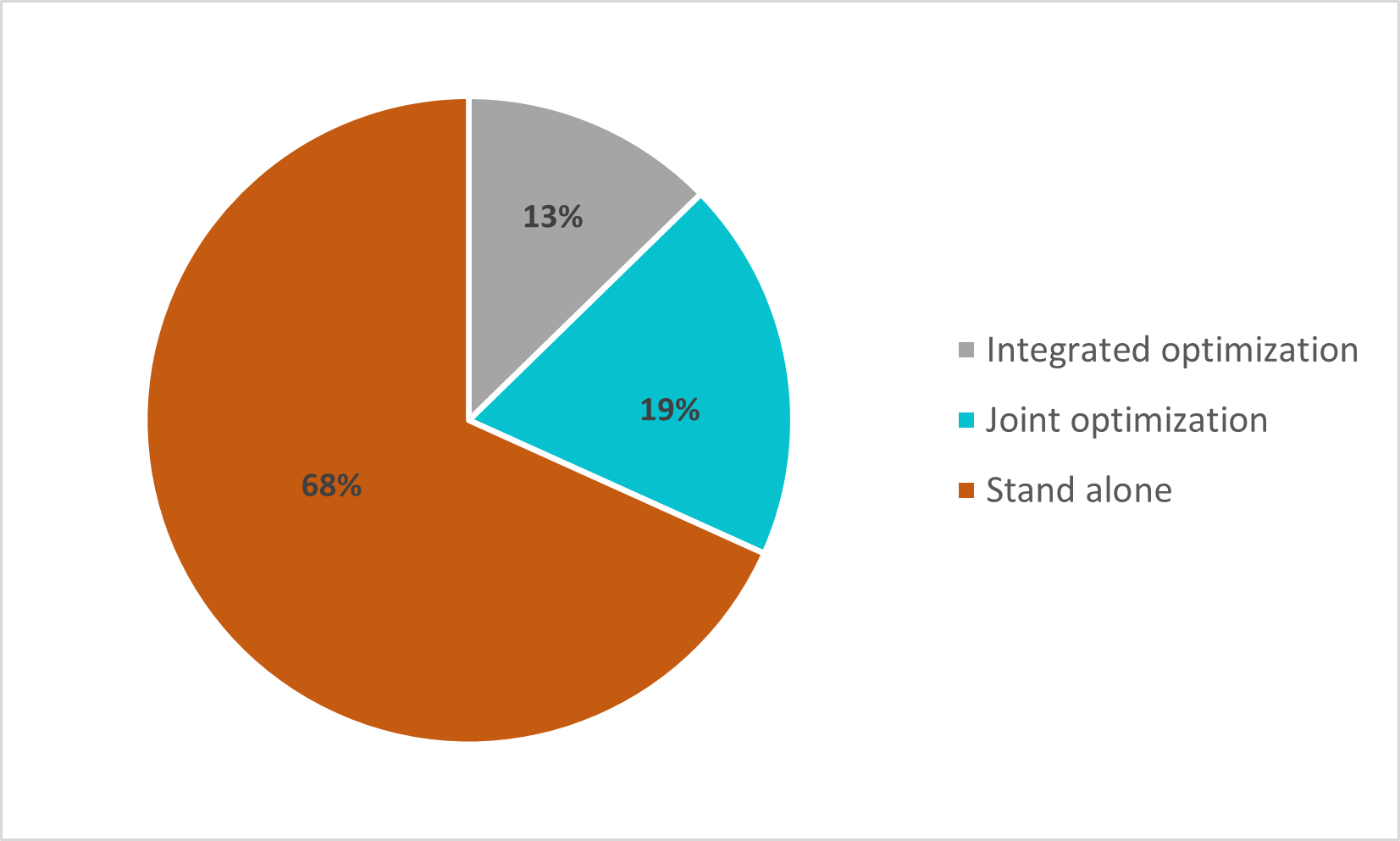}
    \caption{The fraction of RL and DRL-based policies that adopted joint, integrated maintenance policies over stand-alone optimization policies}
    \label{fig 7}
\end{figure}

Maintenance is one of the critical elements of a production system and it can affect productivity significantly. Machine maintenance can affect production yield and product quality, consume production time due to machine unavailability, and unplanned maintenance actions can even disrupt production plans. Conventionally, the entities of production such as production scheduling, inventory, material handling, shift scheduling, and quality assurance and maintenance planning have been treated independently for managing manufacturing systems, while individual planning of these facets of a manufacturing system might achieve the required performance of that section, due to the connected and sometimes conflicting relationships between all these facets of production, the overall system performance might not be achieved. From a managerial point of view, according to \cite{r48}, the integrated, overall optimal performance of the overall production or manufacturing system is more important than individual sections of the production system doing well and conflicting with each other, individual optimization hinders efficiency and smart manufacturing involves combining the individual components of manufacturing into an integrated platform. \\
To efficiently coordinate maintenance activities with other activities, joint and integrated optimization policies are developed. While integrated and joint optimization is used interchangeably in the literature, it can, however, be observed that the outcome of some of these policies is still to schedule maintenance actions while taking into consideration other entities of production, like production plans or inventory levels and the outcome of some other policies involves the simultaneous planning of maintenance activities and other activities such as production, resources and spare part allocations. In this work, in order to distinguish these outcomes, the integrated optimization scope refers to when the outcome is to only decide what machines to maintain and when to carry out the maintenance activities while considering other related entities in the environment but for joint optimization, the output is not to solely decide when and what machines to maintain. Other entities of the environment are planned simultaneously with the maintenance actions, for instance, what and when to produce, and the number of spare parts required in inventory can be planned alongside the machine maintenance activities. \\
The authors of \cite{r50} developed a joint optimization policy that simultaneously schedules flow shop production plans with maintenance actions by introducing the machine maintenance constraints into the multi-factory production scheduling problem formulation.  Some papers like  \cite{r51} tried to find a cost-effective implementation of the corrective maintenance by simultaneously optimizing different (but interconnected) planning decisions, in this paper three (3) separate objectives were optimized, the number of repairable spare part stock to be kept in the inventory was optimized, maintenance workforce capacity optimization was the second planning decision and the third objective was to develop a cross-training policy for the workers where each worker can only repair a subset of all part types to minimize downtime with the minimum cost. \cite{r52} developed a joint optimization policy that combines production, maintenance, and quality policies to contribute to the production system’s total cost-effectiveness. The RL agent can choose from the action set \{produce, maintain, remain idle, or recycle second-class product\}.\\ The authors in \cite{r53} also developed an integrated production and maintenance policy that modeled a stochastic production or inventory system that is subject to deterioration failures, and a maintenance policy that tries to find an optimal trade-off between maintaining a high service level of machines and reducing the inventory level as much as possible is developed, the RL agent’s admissible actions were to either produce, maintain, or remain idle. Figure \ref{fig 7} shows the fraction of RL and DRL-based policies that adopted joint, integrated maintenance policies over stand-alone optimization policies.\\
In this section, we have defined, summarized, and classified the factors and elements that should be considered to formulate the first stage of the maintenance optimization problem for RL and DRL-based solutions for maintenance optimization, the second stage involves three (3) major tasks which are: 
(a) Formulating a maintenance problem into the RL framework.
(b) Proposing a reasonable reward.
(c) Implementing the RL algorithm.
In the next section, these tasks will be further discussed, and relevant publications analyzed.

\newpage
\section{The Reinforcement Learning Problem Formulation} \label{sec:graph}

Reinforcement Learning is a data-driven, decision-making algorithm that learns a sequence of optimal decisions by interacting with the environment, to get the algorithm to learn good actions, the reward and penalty theory is adopted. The interaction between the RL algorithm and the environment can either be model-based or model-free, if the dynamics of the system can be adequately represented by a function, and the transition probabilities are known, then the problem can be constructed as a model-based RL approach, but if the dynamics of the system is an unknown function which is most common in real-world applications and production systems where the dynamics of a manufacturing plant is usually uncertain and stochastic, the problem is them constructed as model-free. For model-free RL implementations, a simulator that tries to mimic the behavior of the real environment is used to train the algorithm. Due to the complexity of the production plant dynamics, getting a function that accurately describes the system is very difficult, and model-free RL is used mostly. figure \ref{fig 8} also supports this claim, it shows the fraction of model-based RL algorithms to model-free in the reviewed publications.
The translation of the maintenance planning problem formulation into the RL framework usually involves the definition of the observable states, allowable actions, reward function, transition probabilities (model-based), or development of the simulation environment (model-free) and choosing the RL algorithm. The following sections summarize the trends and classify reviewed publications in terms of the RL problem formulation. 
\subsection{States}
The RL states are a representation of the environment accessible to the RL agent at a given instance, the states are constantly changing based on the interaction of the RL agent with the environment and depending on the nature of the problem or the complexity of the environment the state changes can either be deterministic or stochastic. It is important for the state captured by the agent to have enough information required for the agent to learn good policies. It can be observed from reviewed publications that the degradation state or level of the machines or assets at every inspection time is the primary state information available to the agent, in all published papers, the degradation state of the machine depends on the maintenance policy upon which the degradation model is developed is used as a representation of the state, amongst other information. \\ Age-based policies use the current age of the machine at every inspection time as one of the state’s representations, an instance of this is in \cite{r40} which adopted the age-based maintenance policy, the effective age(s) of the components which is a continuous, uncountable, mixed-integer state space is used to represent the machine states accessible to the agent. \cite{r18}, \cite{r19},\cite{r49}, \cite{r25},\cite{r45}, \cite{r56}, all used the ages of the machines or assets as a representation of the states amongst other information such as the buffer levels, machine status, and remaining maintenance duration. For failure-limit policies, the degradation stages of the machines or assets based on the machine health index parameter chosen are used to represent the states.\\
It can be observed in most publications that to avoid making assumptions about the degradation model and its respective parameters when data is not available, the Markov discrete state degradation model is commonly used to represent the machine or asset degradation level, for instance, \cite{r4} used four (4) degradation levels (0, 1,2,3) as the state representation where level 0 is referred to as the best functional state and level 3 the most degraded state, also in \cite{r34}, the pump states were defined to be between any of the four (4) levels, (level 1 = No degradation, level 2 = Moderately degraded, level 3 = Severely degraded and level 4 = Failed state). Also, as observed in \cite{r35} and \cite{r41}, even though degradation models were developed to capture the degradation status of the machines or assets, the continuous degradation levels were discretized, and these discretized states were used as state information for the agent. According to \cite{r35}, since the degradation level of the components is used to make maintenance decisions, for systems deteriorating over time, it is more cost and computationally efficient to implement the maintenance action based on discretized deterioration levels. For each component, the degradation states are classified into multi-stages or levels based on the predefined failure thresholds, the degradation levels were broadly divided into four (4) stages; (Stage 0 = component is new and degradation level is zero, Stage 1 = degradation level of component is between threshold levels H2 and H3, Stage 2 = degradation level of component is between threshold levels H2 and H1 and Stage 4 = Failure stage and degradation level of the component is above the threshold level H1).\\
 To capture the variations in the state representations used in reviewed publications, Table \ref{tab 32} classifies all the publications in terms of the deterioration state space, we distinguish deterioration processes with three or four states, a discrete state space, continuous state space and the adopted maintenance policies.
 
\renewcommand{\arraystretch}{1.5}
\begin{table}[htbp]
  \scriptsize
  \centering
  \caption{Reviewed publications grouped in terms of the deterioration of state space and adopted policies.}
    \begin{tabular}{ p{3.2cm}|p{3.2cm} p{3.2cm} p{3.2cm}  }
    \toprule
    \multicolumn{4}{c}{Adopted Policy} \\
    \midrule
    \multicolumn{1}{l}{State-space representation} & Age-based & Failure-limit & Periodic \\
    \midrule
    Three-state discrete space &    -   &  \cite{r31}, \cite{r43} &  - \\
    Four-state discrete space & \cite{r39}, \cite{r49} & \cite{r57}, \cite{r21}, \cite{r45} &  - \\
    Discrete state space & \cite{r18}, \cite{r25}, \cite{r44} & \cite{r7}, \cite{r27}, \cite{r35}, \cite{r52}, \cite{r3}, \cite{r28}, \cite{r34}, \cite{r22}, \cite{r50}, \cite{r24}, \cite{r58}, \cite{r59}, \cite{r60}, \cite{r43} & \cite{r51} \\
    Continuous state space & \cite{r19}, \cite{r56}  & \cite{r53}, \cite{r26}, \cite{r41}, \cite{r40}, \cite{r63} &  \cite{r23} \\
    \bottomrule
    \end{tabular}%
  \label{tab 32}%
\end{table}%

\subsection{Actions}
Actions are taken by an agent to change the states. For the maintenance planning problem, the allowable actions available to the agent to pick from and the effects are directly related to how the maintenance degrees and effects are modeled in the first stage. Also depending on the optimization scope, the allowable actions can vary greatly, for instance, in stand-alone maintenance planning policies, the actions space usually includes corrective, preventive, or minimal repair actions; maintenance actions basically depend on the maintenance activities that can be carried out in the environment. For integrated and joint optimization, the actions are usually based on multiple decisions such as in \cite{r53}, the allowable actions are either to produce, perform corrective or minimal maintenance or remain idle. 
\subsection{Reward}
The reward function is an incentive mechanism that helps the agent know when it is taking good or bad actions based using rewards and punishments. The agent’s goal is to maximize the total reward, the reward function formulation is a major step in developing an effective RL algorithm. Defining reward functions for complex real-world applications can be very difficult, however, for maintenance planning problems, if the optimality criterion or objective function has been adequately defined, then the reward function is usually the negative of the objective function because it is a maximization problem. For this reason, researchers spend a lot of time trying to carefully develop the mathematical representation of the system dynamics and formulating the right objective functions in the first stage of the problem formulation.

\subsection{The Reinforcement Learning Algorithms}
\begin{figure}[h!]
    \centering
    \includegraphics[width=0.7\textwidth]{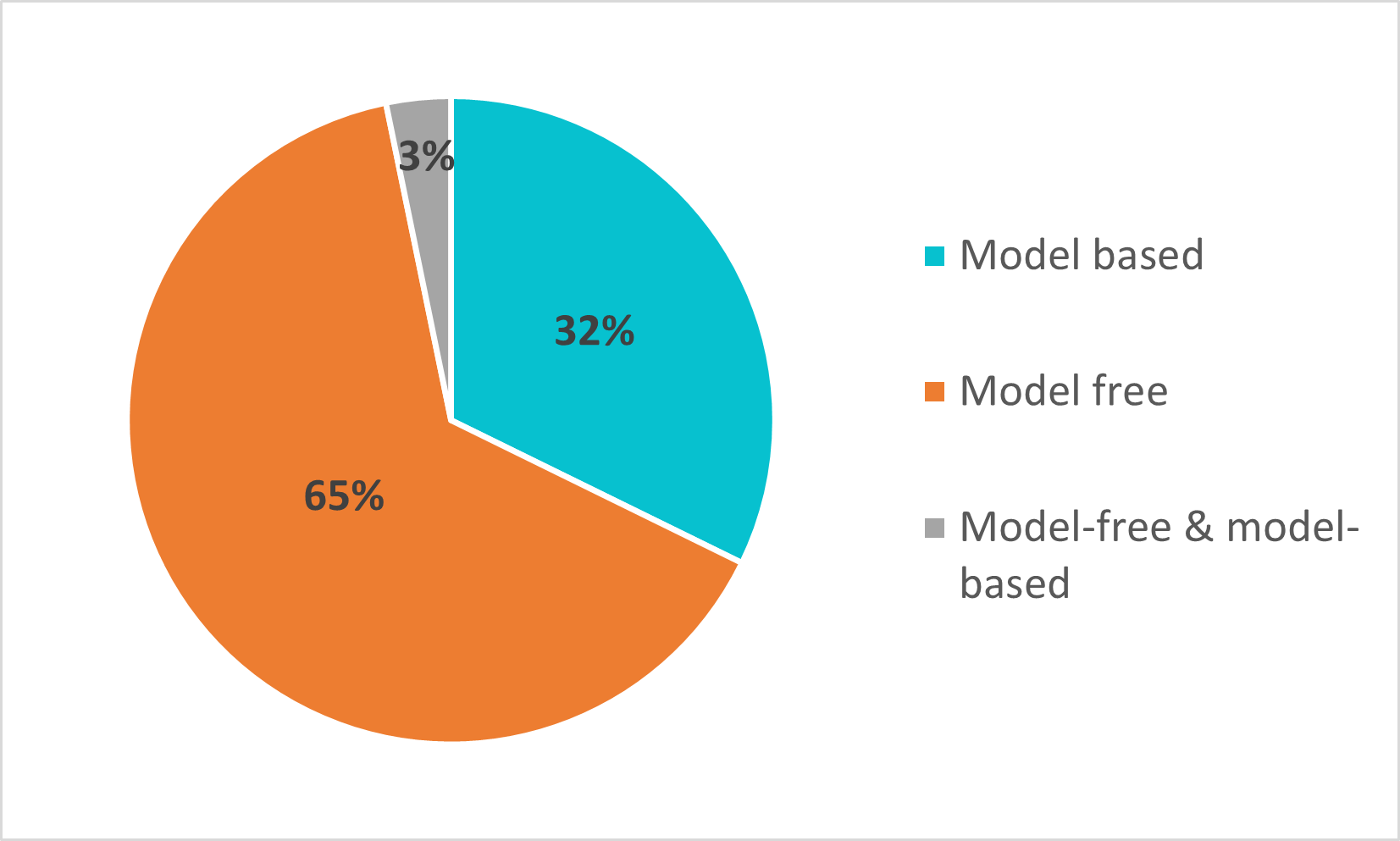}
    \caption{The ratio of Model-based to model-free RL in reviewed publications.}
    \label{fig 8}
\end{figure}

Reinforcement learning algorithms can be classified in many ways: model-based or model-free, classical formulation vs deep learning formulations, single agent vs multi-agents, and based on the extensions of RL algorithms such as RL and meta-heuristics. The publications reference table at the end of this review paper, Table \ref{Reference tables} categorizes each paper in terms of the model type (model-based or model-free), RL agent type (multi or single Agents), and the type of RL algorithms used. It also classifies the papers in terms of stand-alone RL and DRL algorithms and hybrid algorithms which are algorithms that combine RL and DRL with heuristics or meta-heuristics for better convergence and performance. 
Figure \ref{fig 8} shows the ratio of model-based to model-free algorithms, figure \ref{fig 9}, the ratio of a single agent to multi-agent RL and DRL algorithms, and figure \ref{fig 10} shows the ratio of classical RL to deep RL and hybrid RL solutions used in literature within the scope of this work respectively.
Below is a detailed review of publications that adopted multi-agents, RL and meta-heuristics, and other extensions of the RL algorithm to solve the maintenance planning problem in the literature.

\subsubsection{Reinforcement Learning and Multi-agents}
\begin{figure}[h!]
    \centering
    \includegraphics[width=0.7\textwidth]{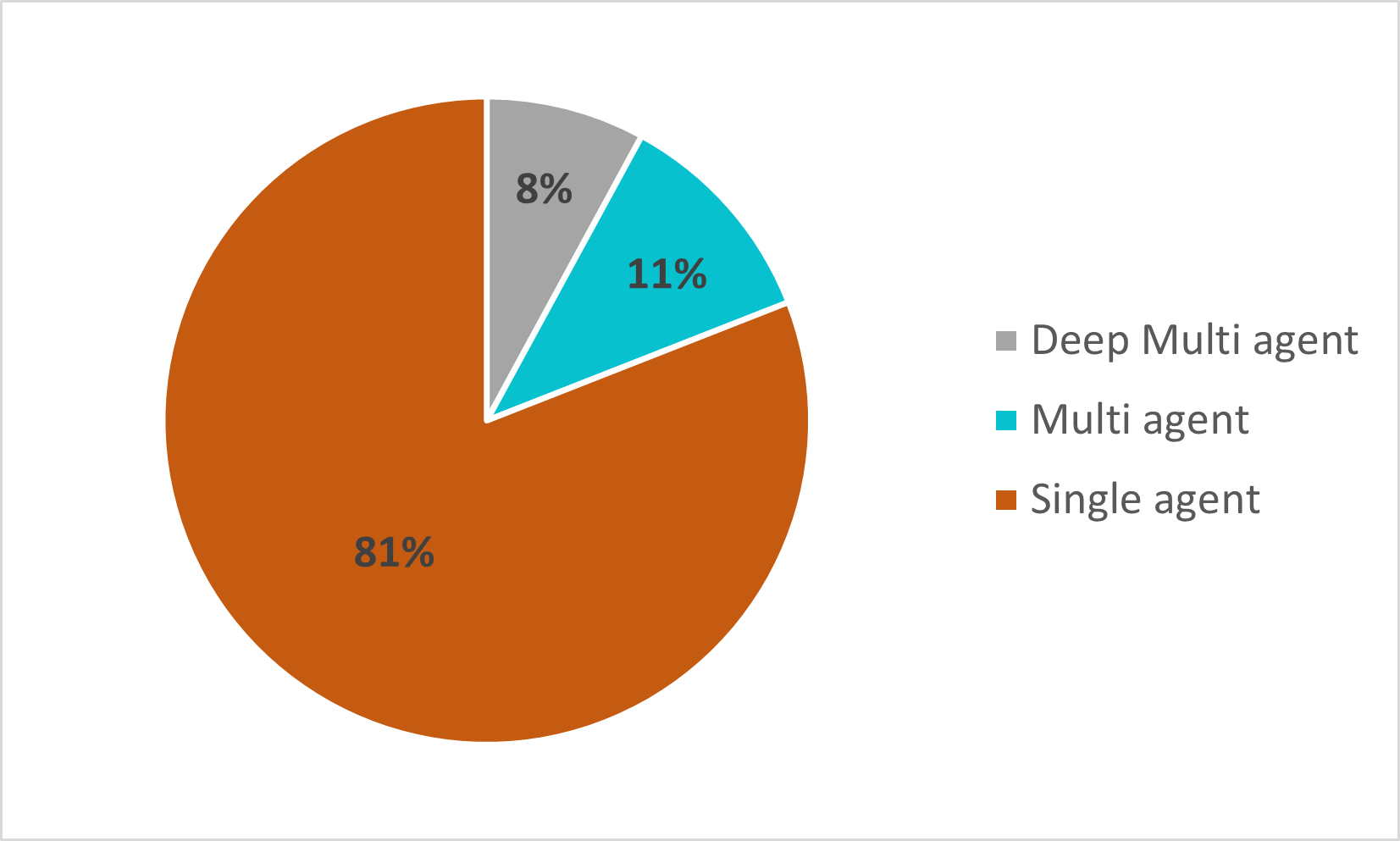}
    \caption{Ratio of Single to Multi-Agent RL in reviewed publications.}
    \label{fig 9}
\end{figure}

The volume of publications that adopted a single-agent RL algorithm is more than those that used multi-agents as seen in figure \ref{fig 9}. The large state and action spaces in maintenance planning problems can pose a challenge to the development of effective maintenance policies, to overcome this difficulty some authors have adopted multi-agent RL algorithms. The authors of \cite{r24} developed a hierarchical-coordinated-multi-agent RL (HMARL)  algorithm to optimize the condition-based maintenance strategy for multi-component systems with large state and action spaces, according to (\cite{r24}, the performance of tabular multi-agent RL (MARL) can be enhanced by using a coordinated mechanism and by arranging the agents in a hierarchical structure hence the adoption of a hierarchically-coordinated-multi-agent-deep Q-learning based RL algorithm to leverage the benefits of the hierarchical structure and the coordinated mechanism in MARL. \\
The authors of \cite{r7} and \cite{r25} also adopted the cost-sharing multi-agent RL algorithm to develop CBM-based maintenance planning policies for serial production lines. \cite{r7} proposed a distributed multi-agent RL algorithm to obtain control-limit maintenance policies for each machine in the flow line system based on the single-agent average reward RL algorithm with the aim of reducing the overall system average cost rate through a learning rule that established a relationship between the local decisions made by each agent depending on the observed state represented by yield level and buffer level of each machine, and the overall optimization goal, this method is referred to as cost-sharing RL algorithm.
 The authors of \cite{r25}, also adopted the cost-sharing multi-agent RL algorithm in combination with heuristics to investigate the maintenance policy for a two-machine-one-buffer (2M1B) assemble line system, and just like in \cite{r7}, the observed states of the deteriorating machines are characterized by the yield level of the machines, the learning efficiency of the RL algorithm in different heuristics search methods was discussed and advantages of heuristic-based RL algorithm was proved.\\
The authors of \cite{r27}, \cite{r40}, and \cite{r62} also used a deep-multi-agent RL algorithm to develop CBM-based maintenance planning policies for multi-unit systems. \cite{r27} proposed optimal maintenance schedules for parallel machines and exploited the unused potentials in maintenance schedules by taking into consideration the opportunistic maintenance policy which aims at determining the point before the breakdown of the machine with low system load and low opportunity cost for maintenance measure rather than waiting for the last possible time. \cite{r40} also proposed a deep-multi-agent-PPO-based RL algorithm approach to developing cost-effective maintenance policies for parallel machines. The authors of  \cite{r62} also proposed a deep-distributed recurrent Q-network multi-agent deep RL algorithm for the optimization of power grid equipment maintenance plans.

\subsubsection{Reinforcement Learning and Meta-heuristics}
In recent years, RL has been integrated with meta-heuristics for combinatorial optimization problems. The maintenance planning problem has also been addressed by some authors in the literature using this approach. The authors of \cite{r25} combined RL with meta-heuristics to improve the performance of meta-heuristics. The authors integrated the artificial bee colony (ABC) algorithm with Q-learning to develop the QABC algorithm used to solve the jointly distributed, three-stage assembly production and maintenance planning optimization problem. The Q-learning algorithm was employed to dynamically select the search operator used by the ABC algorithm. The search operators for ABC algorithms are usually static and adjusted seldomly, however, the adoption of Q-learning allows for dynamic selection of search operators which can improve the exploration ability of the ABC algorithm.
The QABC algorithm was compared with other methods like the ordinary ABC algorithm, Memetic Social Spider optimization algorithm (MSSO), Hybrid Genetic Algorithm-Random Variable Neighbour Search (HGA-RVNS), and Hybrid Genetic Algorithm-Random Variable Neighbour Search (HDDE-RVNS), and the authors were able to show that in terms of performance, the QABC algorithm outperformed the ABC, HGA-RVNS, HDDE-RVNS algorithms on most test instances. When QABC was compared with MSSO, QABC has smaller minimum values than MSSO in some instances however the performance differs greatly as the problem dimension increases. In a case like this, while meta-heuristic is solely used for the maintenance optimization process, the RL algorithm was used to assist the meta-heuristic algorithm to learn better.\\
The authors of \cite{r51} combined DRL with the Simulated Annealing (SA) algorithm, the DRL algorithm was also used to enhance the SA algorithm to solve the maintenance optimization problem. The DRLSA algorithm which is a combination of DRL and SA was developed and, in every training episode rather than generating a new design point or structure randomly in the vicinity of the current design, the initial solution is the best solution found by the DRL algorithm. The best solution from DRL is given to the SA algorithm and the next solution of the SA algorithm is used by the DRL algorithm as the initial state to search for another best solution, this process is repeated continuously in every episode. Through this continuous exchange of information, the DRL algorithm learns to choose the best neighborhood structure to use based on experience gained from past episodes to enhance the performance of the SA algorithm. The proposed DRLSA algorithm was compared with the ordinary SA, Genetic Algorithm, and ML-based clustering algorithms (K-Median 1,2,3 and 4), and the DRLSA algorithm outperformed other algorithms.\\
Also, the authors of \cite{r70} also adopted a combination of multi-agent RL with a Genetic algorithm (GA) to optimize the maintenance decision-making for multi-component systems. As the number of agents increases, the reward function formulation process becomes complicated which makes it difficult for multi-agent RL to converge to the optimal strategy. To achieve better convergence, the genetic algorithm was used as the central unit to guide the decisions of each agent and establish a bilateral interaction mechanism between multi-agent RL and the genetic algorithm. The authors reported that their proposed algorithm is superior in terms of the quality of the solution obtained to the GA and multi-agent RL algorithms used separately.\\
While the integration of meta-heuristics with RL according to the authors are achieving better results compared to the ordinary meta-heuristics algorithms (not integrated with RL), it can be observed that the Q-learning or Deep Q-networks are the algorithms that have been used to assist the meta-heuristics algorithm for maintenance optimization problems, an extension to the existing RL-meta-heuristics algorithms worth exploring is to use other RL algorithms such as SARSA or the policy-based algorithms in combination with meta-heuristics.

\subsubsection{Reinforcement Learning and Other Methods}
Heuristically accelerated-RL methods have also been adopted in literature to optimize the maintenance decision-making process. The authors of \cite{r25} proposed a maintenance policy for a two-machine-one-buffer (2M1B) assembly line system by using a heuristically accelerated multi-agent-RL (HAMRL) method. The cost-sharing-multi-agent-RL algorithm was enhanced with heuristics to speed up the learning process. This method was compared with simulated annealing search (SAS) and neighborhood search (NS) in RL.\\
The authors of \cite{r68} also developed a joint optimization policy for preventive maintenance and production scheduling in multi-state production systems using a heuristically enhanced R-learning RL algorithm. This method is referred to as the HR-learning algorithm and it was compared with the GR-learning algorithm. The R-learning algorithm is a variant of Q-learning The authors of \cite{r68} for infinite-horizon average reward setting. The GR-learning algorithm is a model-free RL algorithm just like R-learning and SMART-learning algorithms that have been proposed for average rewards. \cite{r68} show that the HR-learning converges faster than R and GR-learning.\\
A rare instance of RL and transfer learning was also seen in literature and it was adopted by \cite{r47}. A predictive maintenance model (PdM) which aims to jointly optimize the machine network uptime and the allocation of human-based resources in an Industrial IoT (IIoT)-enable manufacturing environment was proposed. The proposed model uses transfer learning (TL) to assist the model-free deep RL algorithm in learning more efficiently by providing it with a significant amount of training data acquired by incorporating expert demonstrations. The proposed method was termed transfer learning with demonstrations (TLDs) by the authors and had a 58\% decrease in training time compared to the baseline methods which do not employ transfer learning.\\
To address the global optimization for infinite-horizon RL-based problems, \cite{r58} developed a linear-programming enhanced RL algorithm for maintenance optimization. The LPRT (linear-programming-enhanced-Rollout algorithm was proposed, the rollout algorithm is a function approximator of the Q-values. The proposed LPRT algorithm results are compared with the results obtained from Linear Programming (LP), multi-agent rollout, GA, and PPO algorithms and results show that the LPRT is able to determine optimal maintenance policies with similar accuracy just as these methods and it is suitable for infinite-horizon problems. 

\begin{figure}
    \centering
    \includegraphics[width=0.7\textwidth]{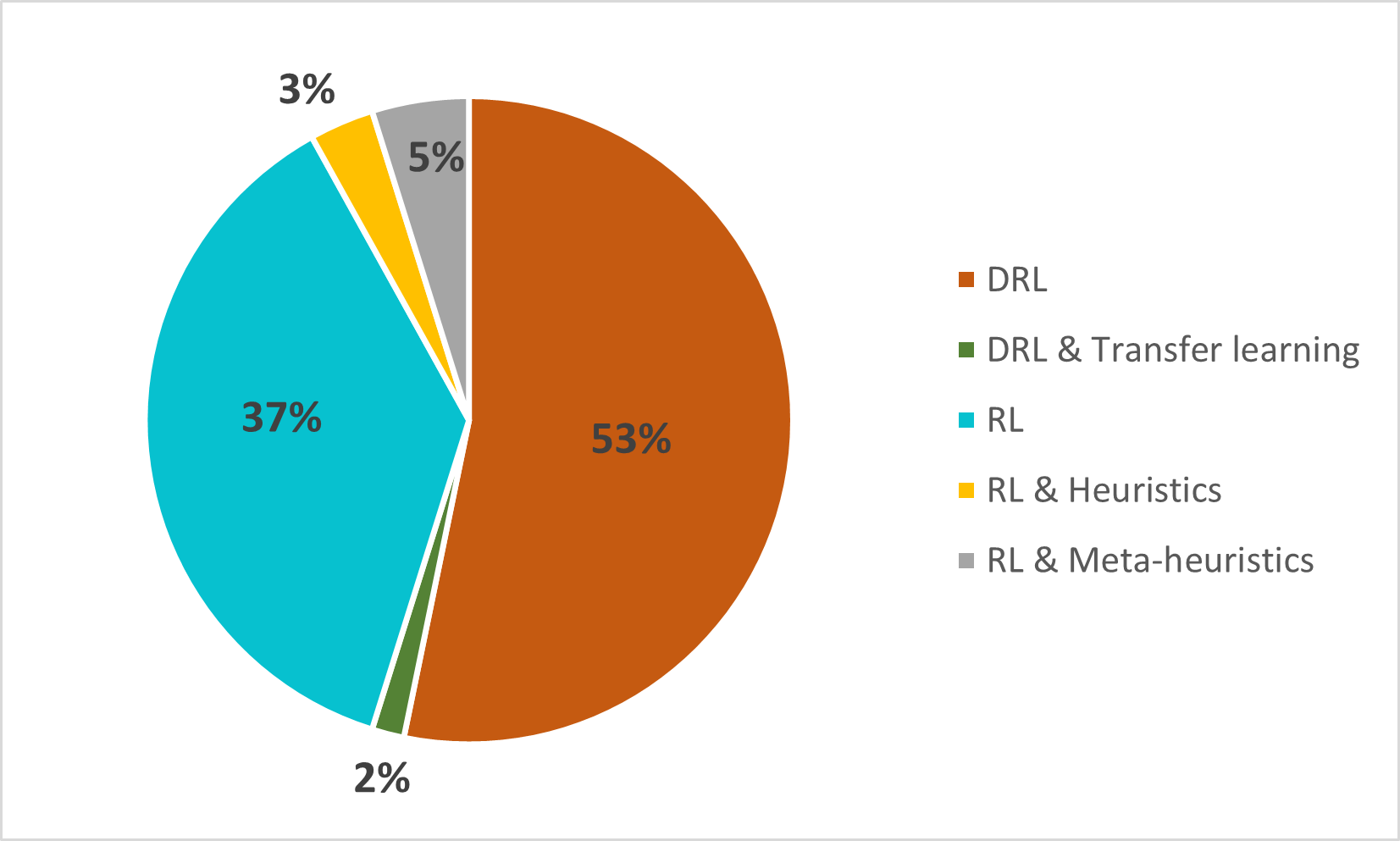}
    \caption{Percentage of reviewed publications with RL, DRL, and Extensions}
    \label{fig 10}
\end{figure}

\begin{figure}
    \centering
    \includegraphics[width=0.7\textwidth]{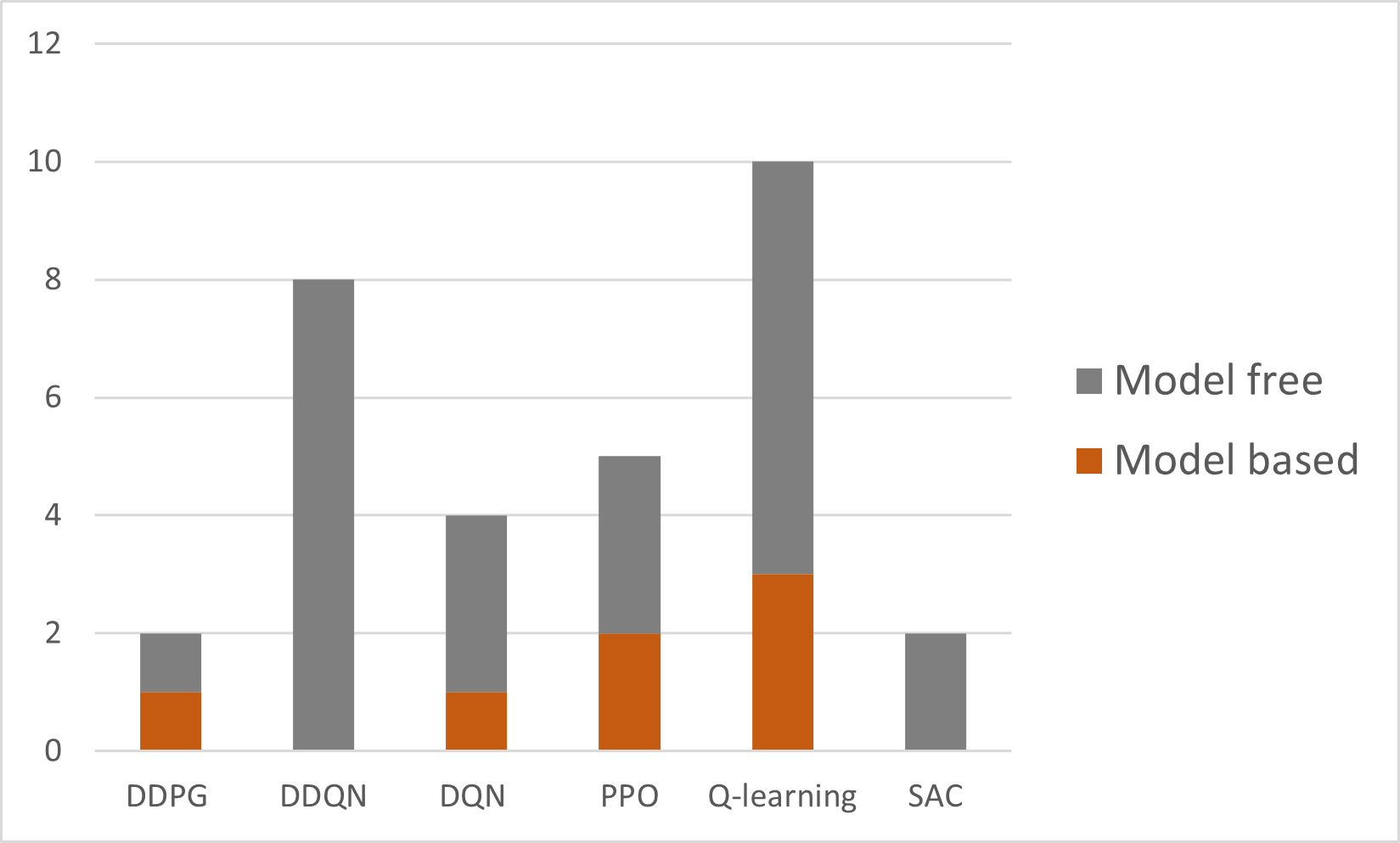}
    \caption{Most used RL and DRL Algorithms in relevant publications}
    \label{fig 11}
\end{figure}

\subsection{General Observations and Insights}
\begin{enumerate}
    \item Due to the complexity of most production environments and industries, getting a function that accurately describes the system is very difficult, so model-free RL is used mostly as seen in figure \ref{fig 8}.
    \item The Q-learning and Deep Q-networks (DQN) are the most used RL algorithms in the literature for maintenance planning problems, as seen in figure \ref{fig 11}, over 70\% of the considered papers used the Q-learning and DQN algorithms. This is because most authors adopted a model-free learning approach and Q-learning algorithms are suitable for model-free learning approaches because they rely on real samples from the environment and not the generated predictions from the next states and rewards to alter their behaviors.
    \item It can also be observed that the condition-monitoring-based maintenance strategy is the most used strategy with the RL and DRL-based solutions and the state space always captures the degradation state or level of the machines or assets. This is because RL is a data-driven optimization algorithm used to find optimum policies in dynamic environments, so to adequately leverage this RL feature a condition-based maintenance strategy is more suitable. Also, from past studies, it has been established that CBM strategies are more efficient than corrective or scheduled maintenance strategies.
    \item Some authors,\cite{r31} and \cite{r28} assumed the existence of predictive health monitoring (PHM) capabilities on the machines or assets, and the RUL information was used as one of the states. For systems where the RUL or time-to-failure can be predicted, it can be a piece of very useful information for the RL agent to make better decisions, however, getting the time-to-failure information for many machines or assets can be very difficult so direct sensor information containing the health or reliability index of the units might be more realistic to come by, this is reflected in literature because most papers used the Failure-limit maintenance policy.
    \item Discrete-event-simulation (DES) is a way of simulating queuing problems or sequential actions. The production environment usually follows a sequential process where entities flow through the system, actions are carried out on these entities, and resources are required to carry out those actions. DES can be used to simulate a production environment and a few authors have used DES as their environment simulators. DES for the environment simulation of maintenance planning and optimization problems was first used by \cite{r53} in literature. Other authors like \cite{r24} and \cite{r52} have since then adopted DES as a simulator for their production environments as well as to train the RL agents and develop efficient maintenance policies.
    Other simulators that have been used in literature for maintenance optimization problems include, Pyomo, Simheuristics, and TensorFlow simulation environment, \cite{r58} used the Pyomo simulator, \cite{r22} used the tensor flow simulation environment, and \cite{r50} used a digital-twin based simulator to train the RL agent to learn optimal preventive maintenance policies. 
    
\end{enumerate}

\newpage
\section{Key Insights of Review Analysis, Implementation Details and Challenges, and Areas of Future Work} \label{sec:levels}
In previous sections, an analysis of RL and DRL-based maintenance policies were carried out, in this section, key insights derived from the review of literature, implementation challenges that have been observed, details of how they have been addressed in the literature, and the areas of future work recommended in some publications highlighted.

\subsection{Implementation details and Challenges}
 A trend in the challenges that have been addressed in literature in terms of the implementation of RL and DRL-based solutions are summarised and discussed in this section. These challenges have been grouped into six (6) main ideas and the papers that have tried to solve these problems and how they have gone about solving them are grouped and discussed. Table \ref{tab 33} groups the papers in terms of the challenges addressed in them.
 \begin{enumerate}
     \item \textbf{Large state and action spaces:} Model-based or model-free RL methods can be used to develop efficient maintenance policies. Model-based RL models utilize known transition probabilities between states but due to the size and complexity of the production environment, model-based RL approaches are limited in their applicability to large state-action spaces because it is impractical to obtain the transition probabilities for larger systems, model-free RL methods, however, can cope with the complex, large, uncertain and stochastic behavior of the production system, the explosion of the state space with increased size encourages the use of the model-free approaches which do not require the transition probabilities.\\ Also, it can be observed that in most maintenance models in the literature that have adopted the RL-based solutions, the degradation states of the system have been discretized to cater to the large state space and according to \cite{r21} and \cite{r50}, the discretization of the degradation states can introduce inaccuracies and inefficiency into the system. To cater to these problems, model-free RL methods with deep RL algorithms are used to handle the explosion of the state space resulting from an increase in the problem size, it also caters to the model inaccuracies that come from the discretization of states when RL-based maintenance policies are used. Multi-agent RL algorithms have also been adopted to cater to large-scale problems. 
     \item \textbf{Timely condition-based planning:} In literature, most developed RL and DRL-based maintenance policies plan maintenance actions at the last minute, they are developed under the assumption that all resources such as personnel and spare parts required for maintenance actions are readily available at every inspection time which isn’t always the case in real applications. Last-minute planning isn’t always desirable because it may cause unexpected downtime during operational hours and does not give room for proper planning of the resources and spare parts unless they are included in the optimization problem formulation, this gives the scheduled maintenance strategies an edge over the CBM-based maintenance policies because they have enough time to plan for the maintenance actions, to cater to this \cite{r31} developed a maintenance policy that incorporates a strategic planning window into the CBM-based maintenance planning decision making that allows for timely maintenance decisions to be made.
     \item \textbf{Infinite Horizon:} In literature, most developed RL and DRL-based maintenance policies are developed for finite horizons and stochastic maintenance scheduling with an infinite time horizon has not been adequately studied. Infinite time horizons are essential for infrastructure or equipment subject to very long-term use, according to \cite{r58}, it is difficult to find an optimal solution for stochastic maintenance scheduling under an infinite horizon due to large computational complexities, a linear-programming-enhanced RL was proposed to consider maintenance scheduling under an infinite time horizon.
     \item \textbf{Joint or Integrated Optimization:} As earlier mentioned, Machine maintenance can affect other entities of production such as production scheduling, inventory, material handling, shift scheduling, and quality assurance. These facets of production are influenced by maintenance actions and if they are planned separately, the overall system performance might not be achieved. From a managerial point of view according to \cite{r48}, the integrated, overall optimal performance of the production system or industry is more important than the individual sections doing well and conflicting with each other. Joint and integrated optimization policies have been developed in the literature to cater to these connected and sometimes conflicting entities and relevant papers are shown in table \ref{Reference tables}. 
     \item \textbf{Algorithm-related solutions:} Different algorithms have been developed by researchers through the combination of one or more optimization algorithms to improve convergence, and computational efficiency and achieve good performance in solving maintenance planning and optimization problems. These algorithmic solutions that have been developed in literature can be broadly grouped into 4 categories. a) RL and DRL and meta-heuristics algorithms b) RL and DRL and heuristics c) RL and DRL and transfer learning algorithms d) RL and linear programming. The papers within the scope of this literature review that used these joint algorithmic solutions for the maintenance planning problem are also highlighted in Table \ref{Reference tables}. An extension of the multi-agent algorithm called the hierarchical-coordinated-RL (HCRL) algorithm adopted by papers \cite{r24} and \cite{r25} was also included. From figure \ref{fig 10}, it can be seen that these joint algorithms have been used in only 10\% of the reviewed publications, with stand-alone RL and deep RL-based algorithms making up 37\% and 53\% of the papers respectively.
     \item \textbf{System-structure related:} Due to the structure of the target system, researchers have developed more specific solutions to cater to them. For instance, due to the structure of non-closed interconnected serial production lines which refers to serial production lines that have buffers between alternating machines, \cite{r18} developed a preventive maintenance policy for serial production lines using a data-driven model which captures all the intricate details of the system dynamics to formulate the problem correctly. The authors of \cite{r26} also developed a system-structure-specific maintenance policy for multi-component systems under competing failure risks. 
 \end{enumerate}
 
\renewcommand{\arraystretch}{1.5}
\begin{table}[htbp]
  \small
  \centering
  \caption{Reviewed publications grouped in terms of the above-mentioned problems addressed in them.}
    \begin{tabular}{ p{5cm}|p{5cm} }
    \toprule
    \multicolumn{2}{c}{Problem Categories} \\
    \midrule
    Large state and action space & \cite{r3}, \cite{r19}, \cite{r21}, \cite{r53}, \cite{r26}, \cite{r41}, \cite{r40}, \cite{r63}, \cite{r56}, \cite{r23}, \cite{r57} \\
    Algorithm-related solutions & \cite{r41}, \cite{r49}, \cite{r34}, \cite{r23} \cite{r24}, \cite{r58}, \cite{r26}, \cite{r51}, \cite{r47}, \cite{r68}, \cite{r70}, \cite{r62}, \cite{r7}, \cite{r27}, \cite{r19} \cite{r40}, \cite{r69}, \cite{r25}, \cite{r70} \\
    Timely planning-based solution & \cite{r3} \\
    System-structure related and multi-unit policies & \cite{r31}, \cite{r18}, \cite{r26}, \cite{r56}, \cite{r72}, \cite{r3}, \cite{r60} \\
    Joint or Integrated optimization & \cite{r7}, \cite{r53}, \cite{r52}, \cite{r22}, \cite{r50}, \cite{r49}, \cite{r4}, \cite{r23}, \cite{r51}, \cite{r25}, \cite{r47}, \cite{r45}, \cite{r68}, \cite{r69} \\
    Infinite Horizon & \cite{r58} \\
    \bottomrule
    \end{tabular}%
  \label{tab 33}%
\end{table}%

\subsection{Key Insights of Review Analysis - Limitations of CBM-based policies.}
 \begin{enumerate}
    \item \textbf{Lack of data to develop adequate maintenance models:} An essential part of developing maintenance planning and optimization models is the modelling of the degradation and occurrence of failure in time and data is required to do this. Also, condition-based, online learning and adaptable RL-based maintenance policies require information about the degradation state of the machines or assets in real-time to make decisions. In real applications this data can be gotten from sensors or IoT devices on the machines or through human-based inspections and monitoring.\\ It can be observed from reviewed papers that the failure distribution or degradation paths derived from the analysis of these sensor data are mostly assumed due to lack of data. These assumptions however cannot be used to build maintenance models that would be used in real-life applications. The lack of data can be due to the huge infrastructural costs associated with installing IoT devices on the assets or costs associated with human-based monitoring of the assets that some companies cannot afford, and those that can afford the costs might deem it unprofitable if the benefits of using condition-based maintenance policies cannot be adequately quantified and weighed against the infrastructure cost of embedding IoT devices on all assets.\\ To cater to this, it is advisable to use condition-monitoring-based maintenance policies for only critical machines or assets. 
    \item \textbf{Generic Solutions:} Due to the factors that affect the development of the RL and DRL-based maintenance policies, it can be observed in the literature that it is challenging to develop a generic solution for the maintenance planning problem due to the size, complexity, optimization scope, structure of the system, and other factors affect the maintenance planning and optimization problem formulation. An opportunity for generic solutions, however, is to develop models based on the system structure, for instance, a model developed for two-machine-one-buffer (2M1B) serial production lines can be adapted for longer serial lines like the five-machine-four-buffer (5M4B) production lines while ensuring that the large state space is catered for. Efficient, generic models based on the system structures can be developed and adapted for other applications.
    \item \textbf{Real-life implementation:} Only a few papers in the literature have implemented the developed RL and DRL-based maintenance policies on real production systems, there are challenges that might be difficult to envisage unless these proposed policies are adapted to real production environments. This is an area that should be explored more because it can reveal new opportunities, ideas, and areas of development for maintenance planning and optimization problems using RL and DRL-based solutions.
   \item \textbf{Robust policies:} It is more practical to use the model-free approach for RL and DRL-based maintenance planning and optimization problems in complex environments and this involves developing a simulation of the environment. Because the RL agent interacts continuously with the environment to learn good policies, when the environment changes, the learned policies also need to be updated. It is common practice for the manufacturing industries to add new resources to accommodate their demands so with every infrastructural upgrade, the CBM policies must also be updated, and this can be challenging. 
\end{enumerate}
\subsection{Areas of future research}
\begin{enumerate}
\item \textbf{Integrated or Joint algorithms:} In literature, it has been observed that meta-heuristics have only been combined with Q-learning and deep Q-learning algorithms, an area of future research is to exploit the potential of other RL algorithms like SARSA with meta-heuristics to develop maintenance planning and optimization policies. Also as suggested by \cite{r7} an area of future research is to consider the combination of distributed multi-agent RL algorithms with meta-heuristics and heuristics to increase the convergence rate and speed up the multi-agent RL learning process.
\item \textbf{Interpretability and Explainability:} One of the criticisms of machine learning models is the lack of interpretability. The interpretability of the developed maintenance policies according to \cite{r18} might be one of the prerequisites for the adoption of RL and DRL-based policies on the real production floor.
\item \textbf{} According to \cite{r24}, the hierarchically-coordinated-multi-agent RL algorithm can obtain better results than the DRL algorithms because the coordinated structure of the HCRL is developed based on characteristics of the maintenance optimization problem, so a potential future research direction is customizing DRL according to the characteristics of the maintenance optimization problem for better performance. 
\item \textbf Numerous maintenance planning and optimization policies that use optimization algorithms like exact solutions, heuristics, and meta-heuristics have been developed in the literature, an area of future research that can be explored would be to use of RL and DRL-based algorithms to solve these problems and compare the results.
\item \textbf{} Finally, in the field of maintenance planning and optimization research, there are no literature review papers that focus on the applications of exact solutions and meta-heuristics for maintenance planning and optimization problems. This work has presented a literature review on RL and DRL-based solutions, the review of literature for meta-heuristics and exact methods remains an area of open research that should be addressed.
\end{enumerate}

\section{Conclusion}
This work reviews the application of reinforcement and/or deep RL algorithms for the development of maintenance planning policies. It emphasizes providing a better understanding of the maintenance planning and optimization problem, the development of the two-stage maintenance planning and optimization problem formulation and model development, and the RL and DRL algorithms used in solving the problem.
In this work, a systematic and integrative review was presented. It focused on highlighting, summarising, and classifying reviewed publications in terms of the methodologies adopted. It presents the findings, and well-defined interpretations of the reviewed studies while finding common ideas and concepts using various graphical and tabular representations. The reference table also provides a holistic review of the literature within the scope of this work.\\
Observed methodological problems and research gaps were identified, key insights from existing literature and areas of future work were also presented in this work. This paper was structured in a way that while capturing the common ideas and well-explored practices in RL and DRL-based maintenance planning and optimization solutions, it also shows the distinct approaches adopted in each publication using tables.\\ Finally, the review provides an understanding of the underlying concepts of RL and DRL-based maintenance planning and optimization problems and solutions and references other resources that can help to gain a deeper understanding of these concepts.

\begin{scriptsize}
\begin{center}
    \begin{longtable}{llp{5.00em} p{5.68em} p{6.365em} p{5.18em} p{5.72em} p{7.275em} p{3.91em}}
    \caption{Reference Table containing a summary of all reviewed papers}\\
    \toprule
    \quad\textbf{Ref No} & \textbf{System} & \textbf{Optimization Scope} & \textbf{Optimality Criteria} & \textbf{Degradation Model} & \textbf{Agent type} & \textbf{RL Model type} & \textbf{RL Algorithm} & \textbf{RL and Extensions} \\
    \midrule
    \endfirsthead

    \multicolumn{9}{l}
    {\tablename\ \thetable\ -- \textit{Continued from previous page}}\\
    \toprule
   \quad\textbf{Ref No} & \textbf{System} & \textbf{Optimization Scope} & \textbf{Optimality Criteria} & \textbf{Degradation Model} & \textbf{Agent type} & \textbf{RL Model type} & \textbf{RL Algorithm} & \textbf{RL and Extensions}\\
    \midrule
    \endhead
    \endfoot
    \cite{r7}     & Multi unit & Integrated optimization & Minimize cost & Gamma distribution & Multi agent & Model free & Cost sharing RL Algorithm & RL \\
    \cite{r53}     & Single unit & Integrated optimization & Minimize cost & Exponential distribution & Single agent & Model free & Q learning & RL \\
    \cite{r31}     & Multi unit & Stand alone & Minimize cost & Assumed PHM capabilities (RUL) & Single agent & Model based & DP    & RL \\
    \cite{r27}     & Multi unit & Stand alone & Minimize cost & Weibull distribution & Deep Multi agent & Model free & DQN, VPG, PPO \& TRPO\newline{} & DRL \\
    \cite{r18}     & Multi unit & Stand alone & Minimize cost & Data-driven relablilty model & Single agent & Model free & DDQN  & DRL \\
    \cite{r26}     & Multi unit & Stand alone & Minimize cost & Gamma and Poisson distribution & Single agent & Model free & DDQN  & DRL \\
    \cite{r39}     & Multi unit & Stand alone & Minimize cost & Weibull distribution & Single agent & Model free and Model based & DQN \& DP\newline{} & DRL \\
    \cite{r35}     & Multi unit & Stand alone & Minimize cost & Gamma and Poisson distribution & Single agent & Model based & Q learning & RL \\
    \cite{r52} & Single unit & Joint optimization & Minimize cost & Exponential distribution & Single agent & Model free & R-learning and greedy Algorithm & RL \\
    \cite{r57}    & Single unit & Stand alone & Minimize cost & Gaussian distribution & Single agent & Model free & Qlearning (SMART) & RL \\
    \cite{r3}    & Multi unit & Stand alone & Minimize cost & Assumed failure probability & Single agent & Model based & DDMAC & DRL \\
    \cite{r41}    & Single unit & Stand alone & Minimize cost & Weiner process and Prognostics(RUL) & Single agent & Model free and model based & Dyna-Q and Q-MA & DRL \\
    \cite{r28}    & Multi unit & Stand alone & Minimize cost & Exponential distribution & Single agent & Model free & PPO   & DRL \\
    \cite{r19}    & Multi unit & Stand alone & Minimize cost & Weibull distribution & Deep Multi agent & Model based & VDMAC & DRL \\
    \cite{r21}    & Multi unit & Stand alone & Minimize cost & Gamma and Poisson distribution & Single agent & Model free & DQN\newline{} & DRL \\
    \cite{r34}    & Single unit & Stand alone & Minimize makespan & Gamma distribution & Single agent & Model based & PERSEUS & RL \\
    \cite{r22}    & Multi unit & Integrated optimization & Minimize makespan & Markov discretized process & Single agent & Model based & Approximate DP & RL \\
    \cite{r40}    & Multi unit & Stand alone & Minimize makespan & Weibull distribution & Deep Multi agent & Model based & PPO   & DRL \\
    \cite{r50}    & Multi unit & Joint optimization & Minimize tardiness & Randomly triggered & Single agent & Model free & DLQL & DRL \\
    \cite{r49}    & Single unit & Joint optimization & Minimize cost & Linear function & Single agent & Model free & DQN\newline{} & DRL \\
    \cite{r4}    & Single unit & Joint optimization & Minimize cost & Markov discretized process & Single agent & Model based & Q learning & RL \\
    \cite{r23}    & Multi unit & Joint optimization & Minimize cost & Predefined time & Single agent & NA    & Q learning (QABC) & Rl and Meta heuristics \\
    \cite{r51}    & Multi unit & Joint optimization & Minimize cost & Not required & Single agent & Model free & DDQN  & Rl and Meta heuristics \\
    \cite{r24}    & Multi unit & Stand alone & Minimize cost & Gamma distribution & Deep Multi agent & Model free & DQL   & DRL (HCRL) \\
    \cite{r58}    & Multi unit & Stand alone & Minimize cost & Assumed failure probability & Single agent & Model based & LPRT  & RL \\
    \cite{r59}    & Multi unit & Stand alone & Maximize profitability & Assumed PHM capabilities (RUL) & Single agent & Model based & DQN\newline{} & DRL \\
    \cite{r60}    & Multi unit & Stand alone & Maximize system performance capacity & Exponential distribution & Single agent & Model based & DDPG\newline{} & DRL \\
    \cite{r25}    & Multi unit & Integrated optimization & Minimize cost & Markov discretized process & Multi agent & Model based & Cost sharing RL Algorithm & RL and Heuristics  \\
    \cite{r43}    & Multi unit & Stand alone & Minimize cost & Assumed PHM capabilities (RUL) & Single agent & Model free & SARSA & RL \\
    \cite{r61}    & Multi unit & Stand alone & Estimate overall equipment efficiency & Known failure rates & Single agent & Model based & Q learning & RL \\
    \cite{r37}    & Single unit & Stand alone & Minimize cost & Poisson distribution & Single agent & Model free & Q learning & RL \\
    \cite{r62}    & Multi unit & Stand alone & Minimize cost & Reliability model & Deep Multi agent & Model free & DRQN  & DRL \\
    \cite{r46}    & Multi unit & Stand alone & Resource optimization & Equipment degradation behaviour from data & Single agent & Model free & PPO LSTM\newline{} & DRL \\
    \cite{r44}    & Multi unit & Stand alone & Minimize cost & Assumed PHM capabilities (RUL) & Single agent & Model based & PPO   & DRL \\
    \cite{r42}    & Single unit & Stand alone & Minimize cost & Weiner process  & Single agent & Model based & GRL\newline{} & RL \\
    \cite{r47}    & Multi unit & Joint optimization & Maximize availability & Equipment degradation behaviour from data & Single agent & Model based & TLD\newline{} & DRL and Transfer learning \\
    \cite{r45}    & Single unit & Integrated optimization & Minimize cost & Assumed PHM capabilities (RUL) & Single agent & Model based & ELM based Q learning & RL \\
    \cite{r63}    & Multi unit & Stand alone & Maximize profitability & Markov discretized process & Single agent & Model free & Monte carlo based RL\newline{} & RL \\
    \cite{r64}    & Multi unit & Stand alone & Minimize cost & Equipment degradation behaviour from data & Single agent & Model free & DDQN  & DRL \\
    \cite{r65}    & Single unit & Stand alone & Minimize cost & Not specified & Single agent & Model free & DQN\newline{} & DRL \\
    \cite{r66}    & Single unit & Stand alone & Minimize cost & Markov discretized process & Single agent & Model based & Gauss Seidel & DRL \\
    \cite{r67}    & Multi unit & Stand alone & Minimize cost & Capacity degradation model & Single agent & Model free & Monte carlo based DRL\newline{} & DRL \\
    \cite{r68}    & Single unit & Joint optimization & Minimize cost & Markov discretized process & Single agent & Model free & R learning (HR and GR-learning) & RL and Heuristics \\
    \cite{r69}    & Multi unit & Joint optimization & Minimize cost & Markov discretized process & Multi agent & Model free & R Smart Algorithm & RL \\
    \cite{r70}    & Multi unit & Stand alone & Minimize cost & Markov discretized process & Multi agent & Model based & MARL (RelaVal) and GA & RL and Meta Heuristics \\
    \cite{r56}    & Multi unit & Stand alone & Minimize cost & Data driven relablilty model & Single agent & Model free & Q learning & RL \\
    \cite{r71}    & Single unit & Stand alone & Minimize cost & Uniformly distribution \& Linear function & Single agent & Model free & Q learning & RL \\
    \cite{r72}    & Multi unit & Stand alone & Minimize cost & Markov discretized process & Single agent & Model free & Q learning & RL \\
    \cite{r73}    & Multi unit & Stand alone & Minimize cost & Markov discretized process & Single agent & Model free & DDQN  & DRL \\
    \cite{rn1}	& Multi unit	& Stand alone &	Minimize cost &	Predefined discrete state markov process and gamma distribution	& Multi agent	& Model based	& Guided probabilistic RL & DRL \\
    \cite{rn2}	& Single unit	& Joint Optimization & Minimize cost	& Data-driven prognostics algorithm - supervised learning regressor	& Single agent	& Model free & 	PPO	 & DRL \\
    \cite{rn3}	& Multi unit	& Stand alone	& Maximize availability	& Not required &	Multi agent	Model free & DDPG & DRL	\\	
    \cite{rn4}	& Single unit	& Stand alone	& Minimize cost	& Data-driven prognostics algorithm-CNN with Monte Carlo dropout & Single agent	& Model free	& SAC	& DRL\\
    \cite{rn5}	& Multi unit	& Stand alone	& Minimize cost	& Poisson distribution	& Single agent & Model free	& DDQN	& DRL\\
    \cite{rn6}	& Single unit	& Joint Optimization	& Minimize cost	& Markov discretized process	Single agent	& Model free	& Q-learning	& RL\\
    \cite{rn7}	& Multi unit & Integrated Optimization	& Minimize cost	& Weibull distribution	& Single agent	& Model free	& DDQN	& DRL\\
    \cite{rn8}	& Multi unit	& Stand alone	& Resource Optimization	& Random process	& Single agent	& Model free	& DDQN	& DRL\\
    \cite{rn9}	& Multi unit	& Integrated Optimization	& Minimize cost	& Gamma distribution	& Multi agent	& Model free	& SMART	& RL\\
    \cite{rn10}	& Multi unit	& Stand alone	& Minimize cost	& Weibull and Gamma distribution & Single agent	& Model free	& PPO	& DRL\\
    \cite{rn11}	& Multi unit	& Stand alone	& Maximize availability	& Weibull distribution	& Single agent	& Model free	& Q-learning	& RL\\
    \cite{rn12}	& Single unit	& Integrated Optimization	& Maximize availability	& Not required &	Single agent	& Model free	& SAC	& DRL\\
    \cite{rn13}	& Multi unit	& Joint Optimization	& Minimize cost	& RUL prediction	& Single agent	& Model free	& Not stated	& Not stated	
    \end{longtable}
    \label{Reference tables}
\end{center}
\end{scriptsize}

\newpage
\section*{Declaration of Conflicting Interest}
The authors declare that they have no known conflicting financial and personal relationships that have influenced the work reported in this paper.

\section*{Acknowledgement}
We would like to acknowledge the financial support of NTWIST Inc. and Natural Sciences and Engineering Research Council (NSERC) Canada under the Alliance Grant ALLRP 555220 – 20, and research collaboration of NTWIST Inc. from Canada, Fraunhofer IEM, D\"{u}spohl Gmbh, and Encoway Gmbh from Germany in this research.

\printcredits
\newpage
\bibliographystyle{ieeetr}

\bibliography{cas-refs}

\end{document}